\documentclass[letterpaper, 10 pt, conference]{ieeeconf}  

\IEEEoverridecommandlockouts                              

\usepackage[export]{adjustbox}

\usepackage{algorithm}
\usepackage[noend]{algpseudocode}
\usepackage{amsmath} 
\usepackage{amssymb}  
\usepackage{tikz}
\usepackage[caption=false]{subfig}
\usepackage{color}
\usepackage{tikz}
\def\etal{\emph{et al}}

\usepackage{algorithm}
\usepackage[noend]{algpseudocode}
\usepackage{tikz}

\usepackage{listings}
\usepackage{xcolor}
\usepackage{multicol}

\lstdefinestyle{customc}{
  belowcaptionskip=\baselineskip,
  breaklines=true,
  frame=L,
  xleftmargin=0.75\parindent,
  language=C++,
  backgroundcolor=\color{black!5}, 
  showstringspaces=false,
  basicstyle=\scriptsize\ttfamily,
  keywordstyle=\bfseries\color{green!40!black},
  commentstyle=\itshape\color{purple!40!black},
  identifierstyle=\color{blue},
  stringstyle=\color{orange},
}

\lstdefinestyle{customasm}{
  belowcaptionskip=1\baselineskip,
  frame=L,
  xleftmargin=\parindent,
  language=[x86masm]Assembler,
  basicstyle=\footnotesize\ttfamily,
  commentstyle=\itshape\color{purple!40!black},
}

\title{\LARGE \bf
POSEAMM: A Unified Framework for Solving \\ Pose Problems using an Alternating Minimization Method*
}

\author{Jo\~{a}o Campos$^{1}$, Jo\~{a}o R. Cardoso$^{2}$, and Pedro Miraldo$^3$
\thanks{*This work was supported by the portuguese FCT project UID/EEA/50009/2019 and (NetSys PhD program), and grant and grant SFRH/BPD/111495/2015. P. Miraldo was partially supported by the Swedish Foundation for Strategic Research (SSF), through the COIN project.}
\thanks{$^{1}$Jo\~{a}o Campos is with the ISR, Instituto Superior T\'{e}cnico, Univ. Lisboa, Portugal.
E-Mail:~{\tt\small jcampos@isr.tecnico.ulisboa.pt}.}%
\thanks{$^{2}$Jo\~{a}o Cardoso is with the ISEC, Instituto Politecnico de Coimbra, Portugal, and the Institute of Systems and Robotics, University of Coimbra, Portugal. E-Mail:~{\tt\small jocar@isec.pt}.}%
\thanks{$^{3}$P. Miraldo is with the KTH Royal Institute of Technology, Stockholm, Sweden.
E-Mail:~{\tt\small miraldo@kth.se}.}
}

\begin{document}

{

\maketitle
\thispagestyle{empty}
\pagestyle{empty}

\begin{abstract}
Pose estimation is one of the most important problems in computer vision.
It can be divided in two different categories --- absolute and relative --- and may involve two different types of camera models: central and non-central.
State-of-the-art methods have been designed to solve separately these problems.
This paper presents a unified framework that is able to solve any pose problem by alternating optimization techniques between two set of parameters, rotation and translation.
In order to make this possible, it is necessary to define an objective function that captures the problem at hand.
Since the objective function will depend on the rotation and translation it is not possible to solve it as a simple minimization problem.
Hence the use of Alternating Minimization methods, in which the function will be alternatively minimized with respect to the rotation and the translation. 
We show how to use our framework in three distinct pose problems.
Our methods are then benchmarked with both synthetic and real data, showing their better balance between computational time and accuracy.
\end{abstract}

\section{Introduction}
\label{intro}
Camera pose is one of the oldest and more important problems in 3D computer vision and its purpose is to find the transformation between two reference frames.
This problem is important for several applications in robotics and computer vision, ranging from navigation, to localization and mapping, and augmented reality.

Pose problems can be divided in two categories: absolute and relative.
In the absolute pose problem, the goal is to find the transformation parameters (rotation and translation) from the world's to the camera's reference frame, using a given set of correspondences between features in the world and their images.
On the other hand, the relative pose aims at finding the transformation between two camera coordinate systems, from a set of correspondences between projection features and their images.
In addition, cameras can be modeled by the perspective model \cite{hartley01,ma03}, known as central cameras, or by the general camera model \cite{grossberg01,sturm04,miraldo11}, here denoted as non-central cameras.
We have noticed that, in the literature, the four cases mentioned above have been in general treated separately, being  each case solved by a specific method (a scheme of those  specific configurations is shown in Fig.~\ref{fig:configurations_posed}.).
In this paper, we aim at proposing a general framework for solving general pose problems (i.e. absolute/relative using central/non-central cameras).

\begin{figure}[t]
\vspace{+0.15cm}
    \centering
    \begin{tikzpicture}
        \node[draw,dotted]{
        \subfloat[Central]{
            \includegraphics[width=0.097\textwidth]{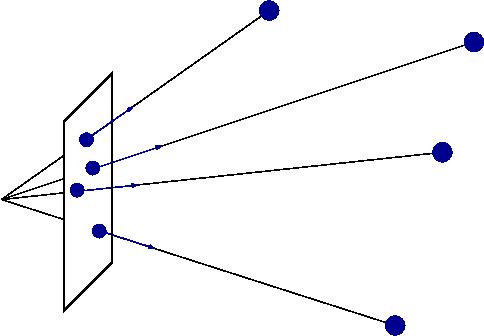}
            \label{fig:apc} 
            } 
        \subfloat[Non-Central]{
            \includegraphics[width=0.097\textwidth]{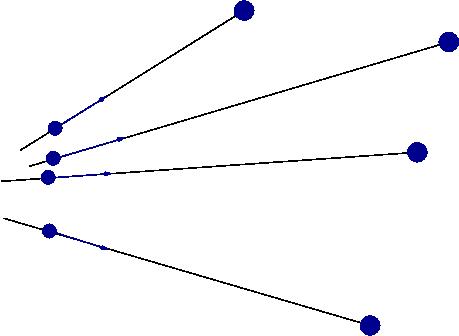}
            \label{fig:apnc}
            }
        };
        \node (text) [anchor=north] at (0,-1.5cm) {\footnotesize \bf Absolute Pose Problems};
    \end{tikzpicture}
    \begin{tikzpicture}
        \node[draw,dotted]{
        \subfloat[Central]{
            \includegraphics[width=0.097\textwidth]{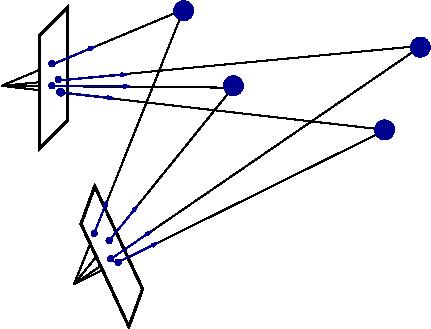}
            \label{fig:rpc}
        } 
        \subfloat[Non-Central]{
            \includegraphics[width=0.097\textwidth]{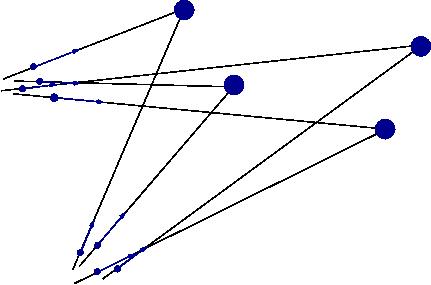}
            \label{fig:rpnc}
        }
        };
        \node (text) [anchor=north] at (0,-1.5) {\footnotesize \bf Relative Pose Problems};
    \end{tikzpicture}
    \caption{Representation of the pose configurations that can be solved using the proposed framework: \protect\subref{fig:apc} \& \protect\subref{fig:apnc} show the absolute pose for central cameras and non-central cameras, respectively; and \protect\subref{fig:apnc} \& \protect\subref{fig:rpc} depict the relative pose for central cameras and non-central cameras, respectively. Blue features (both 3D points and 3D lines in the camera and world coordinates respectively) are the input data.}
    \label{fig:configurations_posed}
\end{figure}

In addition to central/non-central \& absolute/relative cases, pose problems may use minimal or non-minimal data.
While the latter consists in estimating the best rotation and translation parameters that fit a specific pose, the former is important for robust random sample consensus techniques (such as RANSAC~\cite{fishler81}). 
Minimal solutions aim at providing very fast solutions (which are achieved by using the minimal data necessary to compute a solution), and their goal is to obtain a solution that is robust to outliers, rather than giving the best solution for the inliers within the dataset.
This means that, even in an environment with outliers, it is important to run non-minimal techniques after getting the inliers from a RANSAC technique to get the best solution.

Being one of the most studied problems in 3D vision, there are several distinct algorithms in the literature to solve each of the problems.
When considering absolute pose problems (see Figs.~\ref{fig:configurations_posed}\subref{fig:apc} and~\ref{fig:configurations_posed}\subref{fig:apnc} for the case of 3D points and their respective images), there are solutions using minimal data with both points and line correspondences for central cameras  \cite{kneip11,ramalingam11,ke17,wang18}, for non-central cameras  \cite{nister04_2,miraldo14,ventura14,lee16,miraldo18}; using non-minimal data using both points and lines for central cameras  \cite{lepetit09,hesch11,zheng13}, and for non-central cameras,  \cite{sweeney14,kneip14_3,kneip14_1,miraldo14_1,sebastian15,miraldo15_2}.

When considering relative pose problems (see Figs.~\ref{fig:configurations_posed}\subref{fig:rpc} and~\ref{fig:configurations_posed}\subref{fig:rpnc}), there are several solutions for the central camera model, both using minimal data \cite{nister04,stewenius06,li06,kukelova08_2} \& non-minimal data \cite{nister04_3,scaramuza11,fredriksson16}; and for general non-central cameras using minimal data \cite{stewenius05,ventura15} \& non-minimal data (using both points and lines) 
\cite{li08,lee13,kneip16}.
An interesting minimal method that combines both absolute and relative problems was recently proposed in \cite{camposeco18}.

In this paper we are interested in non-minimal solvers, i.e. we assume that, if necessary, a RANSAC technique was used to get the best inliers before applying our method.
We propose a more generic and simpler approach to solve these problems, that can be used in all pose problems.
We managed to do so by formulating our problem as an optimization one.
Thus, it is necessary to provide an expression for the objective function and the gradients for both the translation and rotation parameters.
In the rest of this section we describe the problem, the challenges, and our contributions.
In Sec.~\ref{sec:our} we present our framework and the involved algorithms.
Sec.~\ref{app} shows some applications in which we use the proposed framework, and the experimental results are presented in Sec.~\ref{exp}. Conclusions are drawn in Sec.~\ref{conclusions}.


\subsection{Problem Statement and Challenges}
\label{intro:prob_formulation}
In its simplest and most general form, the solution to any pose problem (whether it is absolute/relative for central/non-central cameras) verifies
\begin{equation}
\label{amm:generic_expression_objective_function}
    \mathcal{F}(\mathbf{R},\mathbf{t}) = \sum_{i=1}^N e_i(\mathbf{R},\mathbf{t},\mathcal{D}_i)^2 = 0,
\end{equation}
where $e_i$ are the geometric/algebraic residuals, $\mathbf{R}\in\mathcal{SO}(3)$ is the rotation matrix, $\mathbf{t}\in\mathbb{R}^3$ is the translation vector, $N$ is the total number of correspondences and $\mathcal{D}_i$ is the known data related with the $i^{th}$ correspondence.
This data may involve correspondences between 3D projection lines (relative pose problems) or between projection lines and 3D points (absolute pose problems).
Under this formulation, any problem can then be stated as
\begin{equation}
\label{eq:problem_def}
\begin{aligned}
& \{\mathbf{R}^*,\ \mathbf{t}^*\} = \underset{\mathbf{R} \in \mathcal{SO}(3),\ \mathbf{t}\in\mathbb{R}^3 }{\text{argmin}}
& & \mathcal{F}\left(\mathbf{R} , \mathbf{t}\right) \\
\end{aligned}.
\end{equation}

In terms of challenges, this problem is, in general, difficult due to its non-linearities:
\begin{enumerate}
    \item $\mathcal{F}$ is usually a high degree polynomial with monomials combining the nine elements of the rotation matrix and the translation vector; and
    \item $\mathbf{R}\in\mathcal{SO}(3)$, i.e. $\mathbf{R}^T\mathbf{R}=\mathbf{I}$, corresponds to nine non-linear  quadratic constraints.
\end{enumerate}

\subsection{Our Contributions and Outline of the Paper}
\label{intro:contributions}
We propose a framework to solve absolute/relative pose problems, for central/non-central camera models, using an Alternating Minimization Method (AMM) and define the corresponding optimization models.
The proposed framework requires as inputs:
\begin{enumerate}
    \item The $\mathcal{D}_i$ upon which the objective function can be obtained;
    \item The specific objective function expression and its Euclidean gradients w.r.t. the rotation ($\nabla g(\mathbf{R})$) and the translation ($\nabla h(\mathbf{t})$).
\end{enumerate} 
Both inputs come from the geometry of the problems and from the derivatives of the objective function.
There is no need for complex simplifications nor additional complex solvers, which have been used to solve this type of problem in the literature.
This is tackled in Sec.~\ref{sec:our}.

To sum up, the main contributions of this paper are:
\begin{enumerate}
    \item The use of an AMM to relax the high degree polynomials associated with $\mathcal{F}$ and their respective constraints (first challenge presented in the previous subsection);
    \item Present steepest descent based algorithms to find the optimal rotation and translation parameters;
    \item Provide some applications of our framework, in which we use the objective functions of \cite{li08}, \cite{gerald08}, and \cite{kneip14_1} (with slight changes to ensure that they do not depend on the number of correspondences).
\end{enumerate}

The proposed technique is evaluated using synthetic and real data, in which we prove that, despite the simple formulation, it significantly improves the computational time when compared with the state-of-the-art techniques.

\section{Solving Pose Problem using Alternating Minimization}
\label{sec:our}
This section presents our generic framework.
We start by describing the Alternative Minimization theory (Sec.~\ref{sec:amm_theory}), and then propose an algorithm to solve a general pose problem (Sec.~\ref{sec:amm_pose}).
Finally, Sec.~\ref{sec:amm_solver} presents the solvers used in the framework.

\subsection{Alternating Minimization Method (AMM)}\label{sec:amm_theory}
The goal of an AMM \cite{csiszar84,niesen04} is to find the minimum of a given objective function depending on two variables $P$ and $Q$, where $P$ belongs to a given set $\mathcal{P}$, and $Q$ to $\mathcal{Q}$. According to \cite{niesen04}, the AMM may be formulated as
\begin{equation}
\{P^*,\ Q^*\} = 
\underset{\left({P} , {Q}\right) \in \left(\mathcal{P} \times \mathcal{Q}\right)}{\text{argmin}}\ \Lambda\left({P} , {Q}\right) ,
\end{equation}
where $\mathcal{P}$ and $\mathcal{Q}$ are the sets of variables, and $\Lambda: \mathcal{P} \times \mathcal{Q} \rightarrow \mathbb{R}$ is the function to be minimized.
The strategy is to fix one of the variables and solve the resulting optimization problem, in an iterative way.
Then, in each iteration, there are two distinct problems that need to be solved:
\begin{align}
\label{eq_solver_1}
{P}_k & =\underset{{P} \in \mathcal{P} }{\text{argmin}}\  \Lambda\left({P} , {Q}_{k-1}\right) \\
\label{eq_solver_2}
{Q}_k & = \underset{{Q} \in  \mathcal{Q} }{\text{argmin}}\ \Lambda\left({P}_{k} , {Q}\right),
\end{align}
starting with a suitable initial guess ${Q}_0$.
The stopping condition for the iterative cycle is
\begin{equation}
    \label{eq:stop}
    |\Lambda\left({P}_k, {Q}_k\right) - \Lambda\left({P}_{k-1}, {Q}_{k-1}\right)| < \tau \ \ \text{or} \ \
    k = k_{\text{max}},
\end{equation}
where $\tau$ is a threshold for the absolute value of the variation of the objective function in two consecutive iterations, and $k_{\text{max}}$ is the maximum number of iterations allowed.

\subsection{AMM for Pose Problems}\label{sec:amm_pose}
Since a pose estimation problem aims at finding a rotation matrix $\mathbf{R}^*$ and a translation vector $\mathbf{t}^*$, the AMM variables ${P}$ and ${Q}$ are set to $\mathbf{R}$ and $\mathbf{t}$, respectively.
In order to use AMM to solve these problems, we need: 1) an expression for the objective function \& its gradients; and 2) solvers to the minimization problems. 

\begin{algorithm}[t]
\vspace{0.2cm}
  \caption{General AMM algorithm for solving pose problems}
  \label{alg:amm}
    {\footnotesize
        \begin{algorithmic}[1]
        \State $\mathbf{t}_0 \gets$ {\tt initial guess};  \algorithmiccomment{Sets an initial guess for the translation}
        \State $\delta \gets 1$; \algorithmiccomment{Defines an initial value for the error}
        \State $k \gets 1$; \algorithmiccomment{Variable identifying the iterations}
        \State $\tau \gets$ {\tt tol}; \algorithmiccomment{Sets the limit tolerance}
        \State $k_{\text{max}} \gets $ {\tt max\_iter}; \algorithmiccomment{Sets the maximum number of iterations}
        \While {$\delta > \tau ~ $ {\bf and} $k < k_{\text{max}}$} \algorithmiccomment{Iterative cicle}
        \State $\mathbf{R}_k \gets \text{argmin}_{\mathbf{R} \in \mathcal{SO}(3)}
 \mathcal{F}\left(\mathbf{R} , \mathbf{t}_{k-1}\right)$; \algorithmiccomment{New rotation}
        \State $\mathbf{t}_k \gets \text{argmin}_{\mathbf{t} \in \mathbb{R}^3}
 \mathcal{F}\left(\mathbf{R}_k , \mathbf{t}\right)$; \algorithmiccomment{New translation}
        \State $\delta = \left|\mathcal{F}\left(\mathbf{R}_k , \mathbf{t}_k\right) - \mathcal{F}\left(\mathbf{R}_{k-1} , \mathbf{t}_{k-1}\right)\right|$; \algorithmiccomment{Updates the error}
         \State $k = k+1$; \algorithmiccomment{Adds one iteration}
        \EndWhile
        \State $\mathbf{R} = \mathbf{R}_k$ and $\mathbf{t} = \mathbf{t}_k$; \algorithmiccomment{Sets the output estimation}
      \end{algorithmic}
  }
\end{algorithm}

Let us consider a generic pose problem, as shown in \eqref{eq:problem_def}.
Depending on the problem, data $\mathcal{D}_i$ can be 2D/2D or 3D/2D correspondences (either relative or absolute poses, respectively).
Then, we can use the method presented in Sec.~\ref{sec:amm_theory} to solve the problem: an iterative method which starts by taking an initial guess on the translation ($\mathbf{t}_0$) and solve for $\mathbf{R}_1$:
\begin{equation}
\label{amm:min_rotation}
\mathbf{R}_1 = \underset{\mathbf{R} \in \mathcal{SO}(3)}{\text{argmin}}\
 \mathcal{F}\left(\mathbf{R} , \mathbf{t}_{0}\right),
\end{equation}
yielding an estimate for the rotation matrix which will be plugged into
\begin{equation}
\label{amm:min_translation}
\mathbf{t}_1 = \underset{\mathbf{t}\in\mathbb{R}^3 }{\text{argmin}}\ \mathcal{F}\left(\mathbf{R}_1 , \mathbf{t}\right).
\end{equation}
This process repeats for all new estimates $\mathbf{t}_k$, until the stopping condition of \eqref{eq:stop} is met.
An overview of the proposed method is given in Algorithm~\ref{alg:amm}.
As a framework, in this stage, one has only to provide $\mathcal{F}\left(\mathbf{R}, \mathbf{t}\right)$, which depends on the specific pose problem to be solved. Below, we present two efficient techniques to solve \eqref{amm:min_rotation} and \eqref{amm:min_translation}.

\subsection{Efficient Solvers for the AMM Sub-Problems \eqref{amm:min_rotation} and \eqref{amm:min_translation}}\label{sec:amm_solver}
To ease the notation, we consider $g\left(\mathbf{R}\right) = \mathcal{F}\left(\mathbf{R}, \mathbf{t}_c\right)$ and $h\left(\mathbf{t}\right) = \mathcal{F}\left(\mathbf{R}_c, \mathbf{t}\right)$, where $\mathbf{R}_c$ and $\mathbf{t}_c$ represent constant rotation and translation parameters.

\begin{algorithm}[t]
\vspace{0.2cm}
  \caption{Generic Solver for the Rotation Matrix. Although $g(\mathbf{R})$ depends on the translation, this variable remains constant during the algorithm, so it is omitted. $\nabla g\left(\mathbf{R}\right)$ stands for the calculated Euclidean gradient of the objective function.}
  \label{alg:ours}
  {\footnotesize
      \begin{algorithmic}[1]
        \State $\mathbf{X}_0\in \mathcal{SO}(3) \gets$ {\tt initial guess};  \algorithmiccomment{Initial guess for the rotation}
        \State $\mu_1 \gets 1$; \algorithmiccomment{Initial angle for the deviation in the $\mathcal{SO}(3)$ manifold}
        \State $\delta \gets 1$; \algorithmiccomment{Sets an initial value for the error}
        \State $\tau \gets$ {\tt tol};  \algorithmiccomment{Sets a limit for the tolerance}
        \State $k \gets 0$; \algorithmiccomment{Initiates the number of iterations}
        \While {$\delta > \tau $} \algorithmiccomment{Optimization cycle}
        \State $\mathbf{Z}_k\gets\nabla g(\mathbf{X}_k)\, \mathbf{X}_k^T-\mathbf{X}_k\nabla g(\mathbf{X}_k)^T$; \algorithmiccomment{Riemannian gradient}
        \State $z_k\gets0.5\ \text{trace}(\mathbf{Z}_k\mathbf{Z}_k^T)$; \algorithmiccomment{Rate at which a rotation step is found}
      \State $\mathbf{P}_k\gets \mathbf{I} + \sin\left(\mu_{k}\right)\mathbf{Z}_{k}^T + \left(1-\cos\left(\mu_k\right)\right) \left(\mathbf{Z}_k^T\right)^2$; \algorithmiccomment{Iterative step}
        \State $\mathbf{Q}_k\gets\mathbf{P}_k\mathbf{P}_k$;  \algorithmiccomment{Initial hypothesis of an iterative step}
        \While{$g(\mathbf{X}_k)-g(\mathbf{Q}_k\mathbf{X}_k)\geq \mu_kz_k$}  \algorithmiccomment{Updates the hypothesis}
        \State $\mathbf{P}_k\gets\mathbf{Q}_k$;  \algorithmiccomment{Updates the iterative step}
        \State $\mathbf{Q}_k\gets\mathbf{P}_k\mathbf{P}_k$;  \algorithmiccomment{Computes the new hypothesis}
        \State $\mu_k\gets2\mu_k$;  \algorithmiccomment{Updates the step}
        \EndWhile
        \While{$g(\mathbf{X}_k)-g(\mathbf{Q}_k\mathbf{X}_k) < 0.5\mu_kz_k$} \algorithmiccomment{Updates the step}
        \State $\mathbf{P}_k\gets \mathbf{I} + \sin\left(\mu_{k}\right)\mathbf{Z}_{k}^T + 
    \left(1-\cos\left(\mu_k\right)\right) \left(\mathbf{Z}_k^T\right)^2$;  \algorithmiccomment{Step}
        \State $\mu_k\gets0.5 \mu_k$; \algorithmiccomment{Updates the rotation angle}
        \EndWhile
        \State $\mathbf{X}_{k+1}\gets\mathbf{P}_k\mathbf{X}_k$; \algorithmiccomment{Computes the new estimate}
        \State $\delta \gets \|\mathbf{X}_{k+1}-\mathbf{X}_k\|_{\text{frob}}$; \algorithmiccomment{Sets the new error}
        \State $k\gets k+1$; \algorithmiccomment{Updates the iterative counter}
        \EndWhile
        \State $\mathbf{R}\gets\mathbf{X}_{k}$; \algorithmiccomment{Returns the best estimate}
      \end{algorithmic}
  }
\end{algorithm}

\vspace{0.25cm}\noindent{\bf Efficient solution to \eqref{amm:min_rotation}:~}We use a steepest descent algorithm for unitary matrices \cite{fiesler96,abrudan08} that does not consider the unitary constraints explicitly. This is achieved by iterating in the $\mathcal{SO}(3)$ manifold.  At the beginning of each iteration, we compute the Riemannian gradient $\left(\mathbf{Z}_k\right)$, which is a skew-symmetric matrix. Geometrically, it corresponds to the axis from which a rotation step will be calculated. Then, we find the angle that, together with the axis, defines the rotation step that will be applied to the rotation at the beginning of the iteration to reduce the value of the objective function. The details are described in Algorithm~\ref{alg:ours}.

\vspace{0.2cm}\noindent{\bf Efficient solution to \eqref{amm:min_translation}:~}
We use another algorithm of steepest descent type \cite{fiesler96}.
In each iteration, the translation gradient is calculated and multiplied by a coefficient. Then it is added to the current translation.
In this way, the solver will converge to a translation vector that will minimize the function for a certain rotation matrix.
Details are shown in Algorithm~\ref{alg:ours_translation}.

\begin{algorithm}[t]
\vspace{0.2cm}
  \caption{Generic Solver for the translation vector. $\nabla h\left(\mathbf{t}\right)$ represents the calculated gradient of the objective function in order to the translation's elements}
  \label{alg:ours_translation}
    {\footnotesize 
      \begin{algorithmic}[1]
        \State $\mathbf{x}_0\in \mathbb{R}^3 \gets$ {\tt initial guess};  \algorithmiccomment{Initial guess for the translation}
        \State $\delta \gets 1$; \algorithmiccomment{Sets an initial value for the error}
        \State $\alpha \gets $ {\tt step}; \algorithmiccomment{Chooses a step}
        \State $\tau \gets$ {\tt tol};  \algorithmiccomment{Sets a limit for the tolerance}
        \State $k \gets 0$; \algorithmiccomment{Initiates the number of iterations}
        \While {$\delta > \tau $} \algorithmiccomment{Optimization cycle}
        \State $\mathbf{x}_{k+1} \gets \mathbf{x}_{k} - \alpha\ \nabla h\left(\mathbf{x}_{k}\right) $; \algorithmiccomment{Updates the guess}
        \State $\alpha = \frac{\left(\mathbf{x}_{k+1} - \mathbf{x}_{k}\right)^{T} \left(\nabla h\left(\mathbf{x}_{k+1}\right) - \nabla h\left(\mathbf{x}_{k}\right)\right)}{\left|\nabla h\left(\mathbf{x}_{k+1}\right) - \nabla h\left(\mathbf{x}_{k}\right)\right|^{2}}$;\algorithmiccomment{Updates $\alpha$}
        \If{ $h\left(\mathbf{x}_{k+1}\right) > h\left(\mathbf{x}_{k}\right) $} \algorithmiccomment{Checks if function value increased}
        \State {\bf break}; \algorithmiccomment{If it is, stop the cycle}
        \EndIf
        \State $\delta \gets \left\|h\left(\mathbf{x}_{k+1}\right) - h\left(\mathbf{x}_{k}\right)\right\|_{\text{frob}}$; \algorithmiccomment{Updates the error}
        \State $k\gets k+1$; \algorithmiccomment{Updates the iterative counter}
        \EndWhile
        \State $\mathbf{t}\gets\mathbf{x}_{k}$; \algorithmiccomment{Returns the best estimate}
      \end{algorithmic}
  }
\end{algorithm}

\vspace{0.25cm}
Keep in mind that Algorithms~\ref{alg:amm}, \ref{alg:ours}, and~\ref{alg:ours_translation}, upon which our general framework is based, only require the objective function $\mathcal{F}(\mathbf{R},\mathbf{t})$ (which depends on the pose problem) and its gradients $\nabla g\left(\mathbf{R}\right)$ \& $\nabla h\left(\mathbf{t}\right)$.

To evidence the simplicity of our framework in solving pose problems, we present, in the next section, three different applications, i.e. we explain how the framework is applied to three different $\mathcal{F}(\mathbf{R},\mathbf{t})$.

Our framework was implemented in {\tt C++} in the {\tt OpenGV} framework. The code are available in the author's webpage.

\section{Applications of Our Framework}
\label{app}
This section presents three applications of the proposed
framework to solve: a relative pose problem (Sec. III-A) and
two absolute pose problems (Secs. III-B and III-C).
\subsection{General Relative Pose Problem}
\label{app:rel_pose_func}
A relative pose problem consists in estimating the rotation and translation parameters, which ensure the intersection of 3D projection rays from a pair of cameras.
Formally, using the {\it Generalized Epipolar} constraint \cite{pless03}, for a set of $N$ correspondences between left and right inverse projection rays (which sets up $\mathcal{D}_i$), we can define the objective function as
\begin{multline}
\label{eq:grel_pose_fo}
\mathcal{F}(\mathbf{R}, \mathbf{t}) = \sum_{i=1}^{N}
\mathbf{v}^{T}\left(\mathbf{a}_{i} \mathbf{a}_{i}^{T} \right)\mathbf{v}  \ \text{with}\ \mathbf{v} = \begin{bmatrix}\mathbf{e} \\ \mathbf{r}\end{bmatrix} \ \  \Rightarrow \\ \ \mathcal{F}(\mathbf{R}, \mathbf{t}) = \mathbf{v}^{T}\mathbf{M}\mathbf{v},\ \text{where} \ \mathbf{M} = \sum_{i=1}^{N}\mathbf{a}_{i} \mathbf{a}_{i}^{T},
\end{multline}
where $\mathbf{a}_i$ is a $18\times 1$ vector that depends on $\mathcal{D}_i$ and $\mathbf{e}$ \& $\mathbf{r}$ are $9\times 1$ vectors built from the stacked columns of the essential \cite{Hartley00} and the rotation matrices, respectively.

The expressions of $\nabla g(\mathbf{R})$ and $\nabla h(\mathbf{t})$ are computed directly from \eqref{eq:grel_pose_fo}:
\begin{equation}
    \nabla g(\mathbf{R}) = 2 
    \frac{d\mathbf{v}^{T}}{d\mathbf{r}}\mathbf{M} \mathbf{v} \ \ \text{ and  } \ \
    \nabla h(\mathbf{t}) = 2 
    \frac{d\mathbf{v}^{T}}{d\mathbf{t}}\mathbf{M} \mathbf{v},
\end{equation}
where $\frac{d\mathbf{v}^{T}}{d\mathbf{t}}$ and $\frac{d\mathbf{v}^{T}}{d\mathbf{r}}$ are matrices whose expressions, due to space limitations, are in the supplementary material. Check the author's webpage.

\subsection{General Absolute Pose Problem}
\label{GPNP}
This section addresses the application of the proposed framework to the general absolute pose, i.e. for a set of known correspondences between 3D points and their respective inverse generic projection rays (as presented in \cite{grossberg01,sturm04,miraldo11}), which set up $\mathcal{D}_i$.
We consider the geometric distance between a point in the world and its projection ray presented in \cite{gerald06,gerald08}.
After some simplifications, we get the objective function
\begin{equation}
\label{app:basic_form}
\mathcal{F}(\mathbf{R}, \mathbf{t}) = \mathbf{r}^{T} \mathbf{M}_{rr} \mathbf{r} + \mathbf{v}_{r}^{T} \mathbf{r} + \mathbf{t}^{T} \mathbf{M}_{tr} \mathbf{r} +
\mathbf{t}^{T} \mathbf{M}_{tt} \mathbf{t} + \mathbf{v}_{t}^{T} \mathbf{t} + c,
\end{equation}
where matrices $\mathbf{M}_{rr}$, $\mathbf{M}_{rt}$, $\mathbf{M}_{tt}$, vectors $\mathbf{v}_r$, $\mathbf{v}_t$, and scalar $c$ depend on the data $\mathcal{D}_i$. Again, due to space limitations, these parameters are in the supplementary material.
The gradients are easily obtained \cite{lutkepohl97}:
\begin{align}
\label{app:basic_form_grad}
    \nabla g(\mathbf{R}) = &~ 2 \mathbf{M}_{rr} \mathbf{r} + \mathbf{v}_r + \mathbf{M}_{tr}^{T} \mathbf{t}\ \text{and}\\ \nabla h(\mathbf{t}) = &~ 2 \mathbf{M}_{tt} \mathbf{t} + \mathbf{M}_{tr} \mathbf{r} + \mathbf{v}_t. \label{app:basic_form_grad_2}
\end{align}

Although not necessary for our framework, one important advantage of having \eqref{app:basic_form}, \eqref{app:basic_form_grad}, and \eqref{app:basic_form_grad_2} in this form instead of the more general formulation of \eqref{amm:generic_expression_objective_function}, is that the calculation of the objective function and its gradients will not dependent on the number of points, leading to a complexity $\mathcal{O}(1)$, instead of $\mathcal{O}(N)$.

\subsection{General Absolute Pose Problem using the UPnP Metric}
\label{UPNP}
In Sec.~\ref{GPNP}, the geometric distance is used to derive an objective function.
In the present case we derive a function based on \cite{kneip14_1}\footnote{The well known method denoted as UPnP.}. The starting point is the constraint
\begin{equation}
\label{app:upnp_1st_system}
    \alpha_i \mathbf{v}_i + \mathbf{c}_i = \mathbf{R} \mathbf{p}_i + \mathbf{t}, \ \ \forall i \in [1, N],
\end{equation}
where $\alpha_i$ represents the depth, $\mathbf{c}_i \in \mathbb{R}^{3}$ is a vector from the origin of the camera's reference frame to a ray's point,  $\mathbf{v}_i \in \mathbb{R}^{3}$ represents the ray's direction, and $\mathbf{p}_i$ is a point in the world's reference frame.
Eliminating the depths $\alpha_i$ will result in an objective function which has the same format as \eqref{app:basic_form}, but matrices $\mathbf{M}_{rr}$, $\mathbf{v}_r$, $\mathbf{M}_{tr}$, $\mathbf{M}_{tt}$, $\mathbf{v}_t$, and scalar $c$ do not depend on the data in the same way as in the previous case (details will also be provided in the supplementary material).

\section{Results}
\label{exp}
This section presents several results on the evaluation and validation of the proposed framework, in the pose problems of Sec.~\ref{app}.
The code, developed in {\tt C++} within the {\tt OpenGV} framework \cite{opengv}, will be made public. We start by evaluating the methods using synthetic data (Sec.~\ref{exp:rel_pose}), and conclude this section with the real experimental results (Sec.~\ref{exp:real_data}).

\subsection{Results with Synthetic Data}
\label{exp:rel_pose}
This section aims at evaluating our framework (Sec.~\ref{sec:our}) under the applications presented in Sec.~\ref{app}, using synthetic data.
More specifically, we use the {\tt OpenGV} toolbox\footnote{The state-of-the-art techniques were already available to use.}. Due to space limitation, we refer to \cite{opengv} for the details on the dataset generation.

\begin{figure*}[t]
\vspace{0.2cm}
\centering
\captionsetup[subfloat]{farskip=0pt,captionskip=0pt}
\subfloat[Sensitivity to noise pixels for the general relative pose]{
\includegraphics[width=.73\textwidth,valign=c]{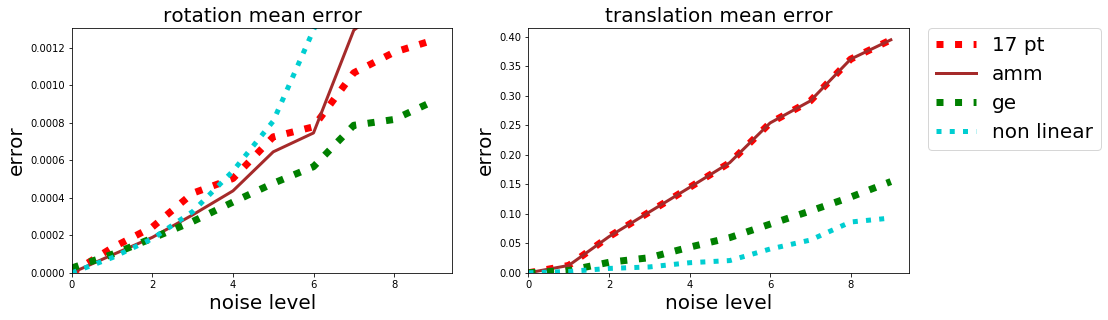} 
\label{fig:rel_noncentral}
}
\subfloat[Execution time (ms) for different algorithms]{
\scalebox{.75}{
\begin{tabular}{ | c | c |}
    \hline 
    {\bf Method} & {\bf Time [ms]} \\
     \hline \hline
    {\tt ge}             & $1.495 \times 10^{ 0}$ \\ \hline
    {\tt 17pt}          & $4.376 \times 10^{ 1}$ \\ \hline
    {\tt amm}            & $1.000 \times 10^{-1}$ \\ \hline
    {\tt non linear}     & $1.223 \times 10^{ 2}$ \\ \hline
   \end{tabular}
   }
   \label{fig:rel_noncentral_ct}
}
\caption{Results for the evaluation of the method proposed in Sec.~\ref{app:rel_pose_func} (which implicitly uses our framework), as a function of the sensitivity to noise \protect\subref{fig:rel_noncentral} and as a function of the required computation time \protect\subref{fig:rel_noncentral_ct}.
The current state-of-the-art techniques were considered.}
\label{fig:relative_pose_results_synthetic}
\end{figure*}

\begin{figure*}[t]
\centering
\captionsetup[subfloat]{farskip=0pt,captionskip=0pt}
\subfloat[Sensitivity to noise pixels for the central absolute pose]{
\includegraphics[width=.73\textwidth,valign=c]{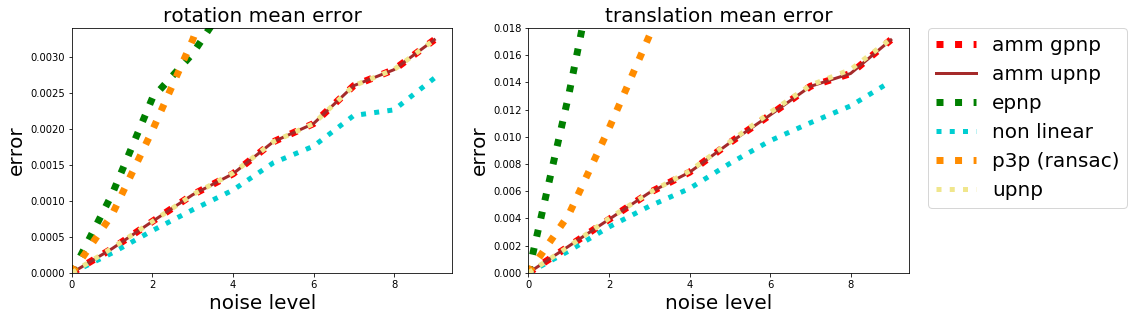}
\label{fig:abs_central}
	}
\subfloat[Execution time (ms) for different algorithms]{
\scalebox{0.75}{
\begin{tabular}{ | c | c |}
    \hline 
    {\bf Method} & {\bf Time [ms]} \\
     \hline \hline
    {\tt epnp}         & $ 8.965 \times 10^{-2}$ \\ \hline
    {\tt upnp}         & $ 5.102 \times 10^{-1}$ \\ \hline
    {\tt non linear}   & $ 4.081 \times 10^{-1}$ \\ \hline
    {\tt amm gpnp}     & $ 6.400 \times 10^{-2}$\\ \hline
    {\tt amm upnp}     & $ 7.152 \times 10^{-2}$ \\ \hline
    {\tt p3p (ransac)} & $ 1.953 \times 10^{-1}$ \\ \hline
   \end{tabular}
   }
   \label{fig:abs_central_ct}
}
\caption{Evaluation of proposed framework with applications of Secs.~\ref{GPNP} and~\ref{UPNP}, in a central absolute camera pose. We evaluate both the errors in terms of noise in the image \protect\subref{fig:abs_central} and in terms of computation time \protect\subref{fig:abs_central_ct}. In terms of state-of-the-art techniques, we consider both the ones with lower computation time and with robust sensitivity to noise.
}
\label{exp:abs_central_info}
\end{figure*}

\begin{figure*}[t]
\centering
\captionsetup[subfloat]{farskip=0pt,captionskip=0pt}
\subfloat[Sensitivity to noise pixels for the non-central absolute pose]{
\includegraphics[width=.73\textwidth,valign=c]{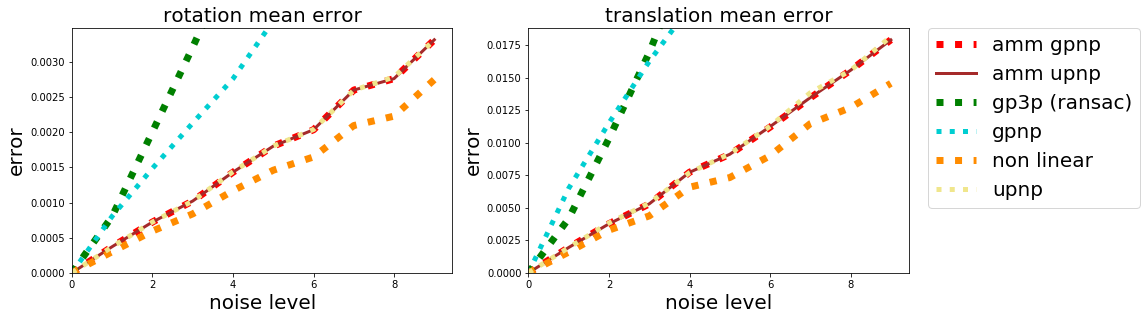}
\label{fig:abs_noncentral}
}
\subfloat[Execution time (ms) for different algorithms]{
\scalebox{0.75}{
\begin{tabular}{ | c | c |}
    \hline 
    {\bf Method} & {\bf Time [ms]} \\
     \hline \hline
    {\tt gpnp}          & $ 3.85 \times 10^{-1}$ \\ \hline
    {\tt upnp}          & $ 5.10 \times 10^{-1}$ \\ \hline
    {\tt non linear}    & $ 4.74 \times 10^{-1}$ \\ \hline
    {\tt amm gpnp}      & $ 6.35 \times 10^{-2}$ \\ \hline
    {\tt amm upnp}      & $ 7.04 \times 10^{-2}$ \\ \hline
    {\tt gp3p (ransac)} & $ 1.38 \times 10^{ 0}$\\ \hline
   \end{tabular}
   }
   \label{fig:abs_noncentral_ct}
}
\caption{Evaluation of proposed framework with applications of Secs.~\ref{GPNP} and~\ref{UPNP}, in a non-central absolute camera pose. We evaluate both the errors in terms of noise in the image \protect\subref{fig:abs_noncentral} and in terms of computation time \protect\subref{fig:abs_noncentral_ct}. In terms of state-of-the-art techniques, we consider both the ones with lower computation time and with robust sensitivity to noise.}
\label{exp:abs_noncentral_info}
\end{figure*}

We start by the relative pose problem addressed in Sec~\ref{app:rel_pose_func}, here denoted as {\tt AMM}, in which the results are presented in Fig.~\ref{fig:relative_pose_results_synthetic}.
We consider the current state-of-the-art techniques: {\tt ge} (Kneip \etal~\cite{kneip14}); the {\tt 17 pt} (Li \etal~\cite{li08}); and the {\tt non-linear} (Kneip \etal~\cite{kneip14_1}).
We use randomly generated data, with noise varying from 0 to 10. For each level of noise, we generate 200 random trials with 20 correspondences between lines in the two camera referential frames, and compute the mean of the errors: 
1) The Frobenius norm of the difference between the ground-truth and estimated rotation matrices; and
2) The norm of the difference between the ground-truth and the estimated translation vectors.
In addition, we store and compute the mean of the computation time required to compute all the 
trials.
The results for the errors are shown in Fig.~\ref{fig:relative_pose_results_synthetic}\subref{fig:rel_noncentral} and for the computation time in Fig.~\ref{fig:relative_pose_results_synthetic}\subref{fig:rel_noncentral_ct}.

Next, we evaluate the techniques in Secs.~\ref{GPNP} and~\ref{UPNP}, for the estimation of the camera pose.
Again, we consider the {\tt OpenGV} toolbox to generate the data, the metrics used were the same as before, as well as the number of trials \& correspondences.
In this case we consider both the central and the non-central cases. Results are shown in Figs~\ref{exp:abs_central_info}\subref{fig:abs_central} and ~\ref{exp:abs_noncentral_info}\subref{fig:abs_noncentral}.
In addition to the methods in Secs.~\ref{GPNP} and~\ref{UPNP} (denoted as {\tt AMM (gpnp)} and {\tt AMM (upnp)}, respectively) for the central case, we consider: {\tt p3p (ransac)} (a closed-form solution using minimal data \cite{kneip11} within the RANSAC framework); {\tt epnp} (presented in \cite{lepetit09}) and {\tt upnp} and {\tt non-linear} (shown in \cite{kneip14_1}) which are state-of-the-art techniques to compute the camera absolute pose.
For the non-central case, we considered: the {\tt gpnp} (presented in \cite{gerald08}); the {\tt upnp} (proposed by Kneip \etal~\cite{kneip14_1}); the  {\tt gp3p (ransac)} (minimal solution \cite{nister04_2} used within the RANSAC framework); and the {\tt non linear} (method presented in \cite{kneip14_1}).

As the initial guess for translation $\mathbf{t}_0$, required by our framework (Algorithm~\ref{alg:amm}), we use the solution given by respective minimal solvers for the absolute pose problems, and the solution given by the linear {\tt 17pt} with a minimum required number of points for the relative pose problem\footnote{Note that this solution with the miminimum number of points is significantly faster than the one shown in Tab.~\ref{fig:relative_pose_results_synthetic}\subref{fig:rel_noncentral_ct} for the {\tt 17pt} that uses all the available points}.
These are very sensitive to noise but very fast, being therefore suitable for a first estimate.

\begin{figure*}[t]
\vspace{0.2cm}
  \centering
  {
  \captionsetup[subfloat]{farskip=1pt,captionskip=1pt}
    \begin{tabular}[b]{c}%
    \subfloat[Used camera system.]{
    \label{fig:real2:setup}
    \includegraphics[height=0.101\textheight]{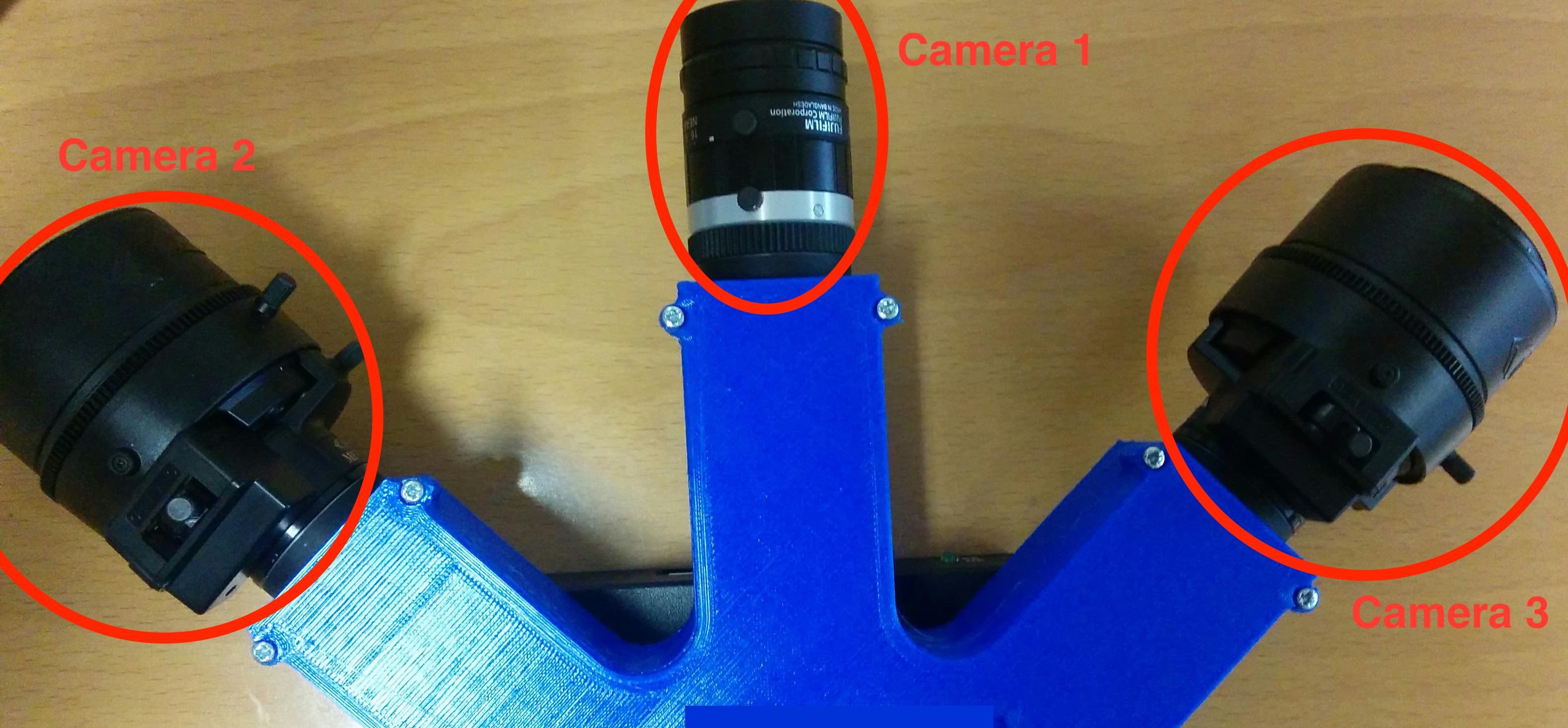}}\\
    \subfloat[Three images acquired at the same instant of time.]{\includegraphics[height=0.053\textheight]{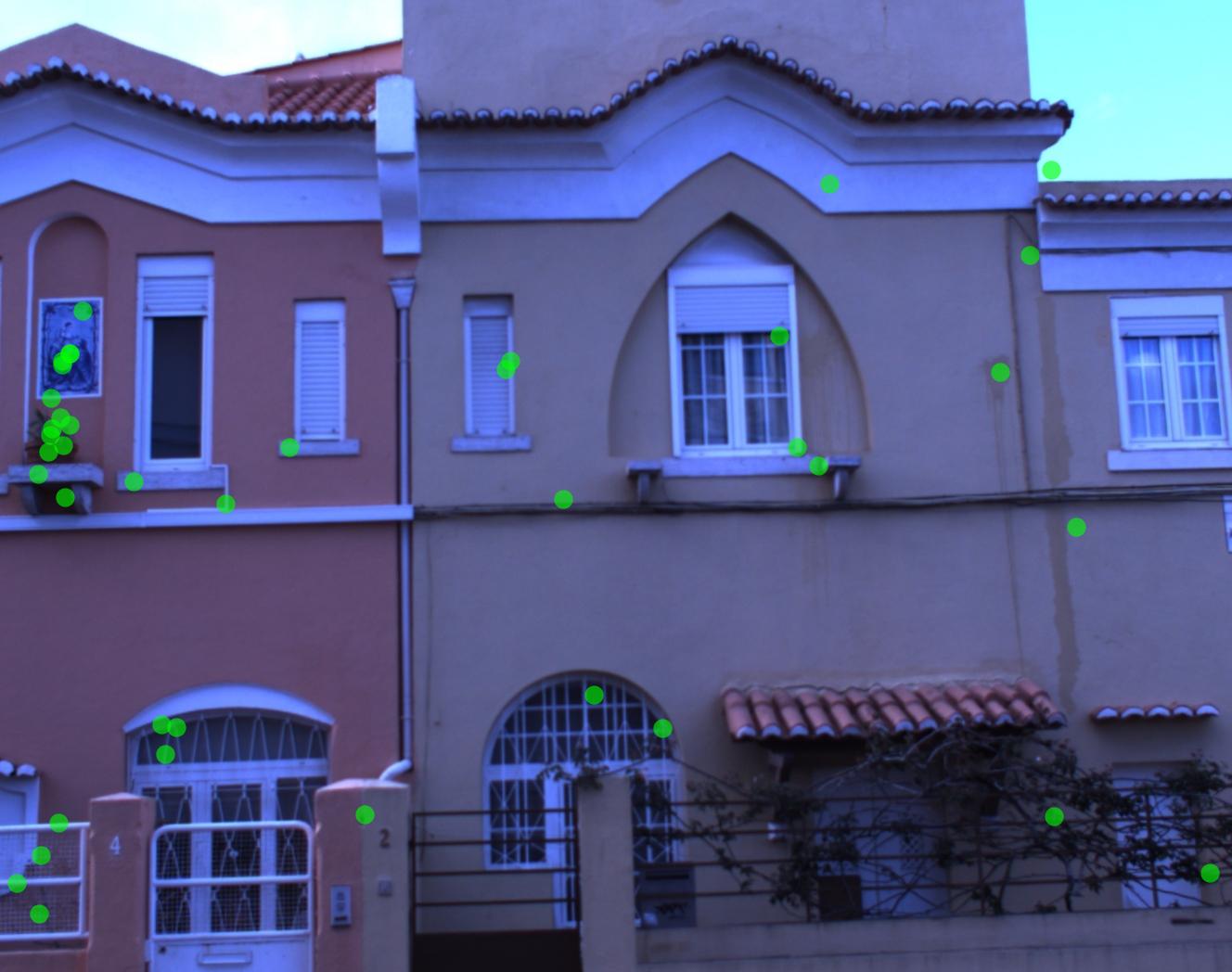}
    \includegraphics[height=0.053\textheight]{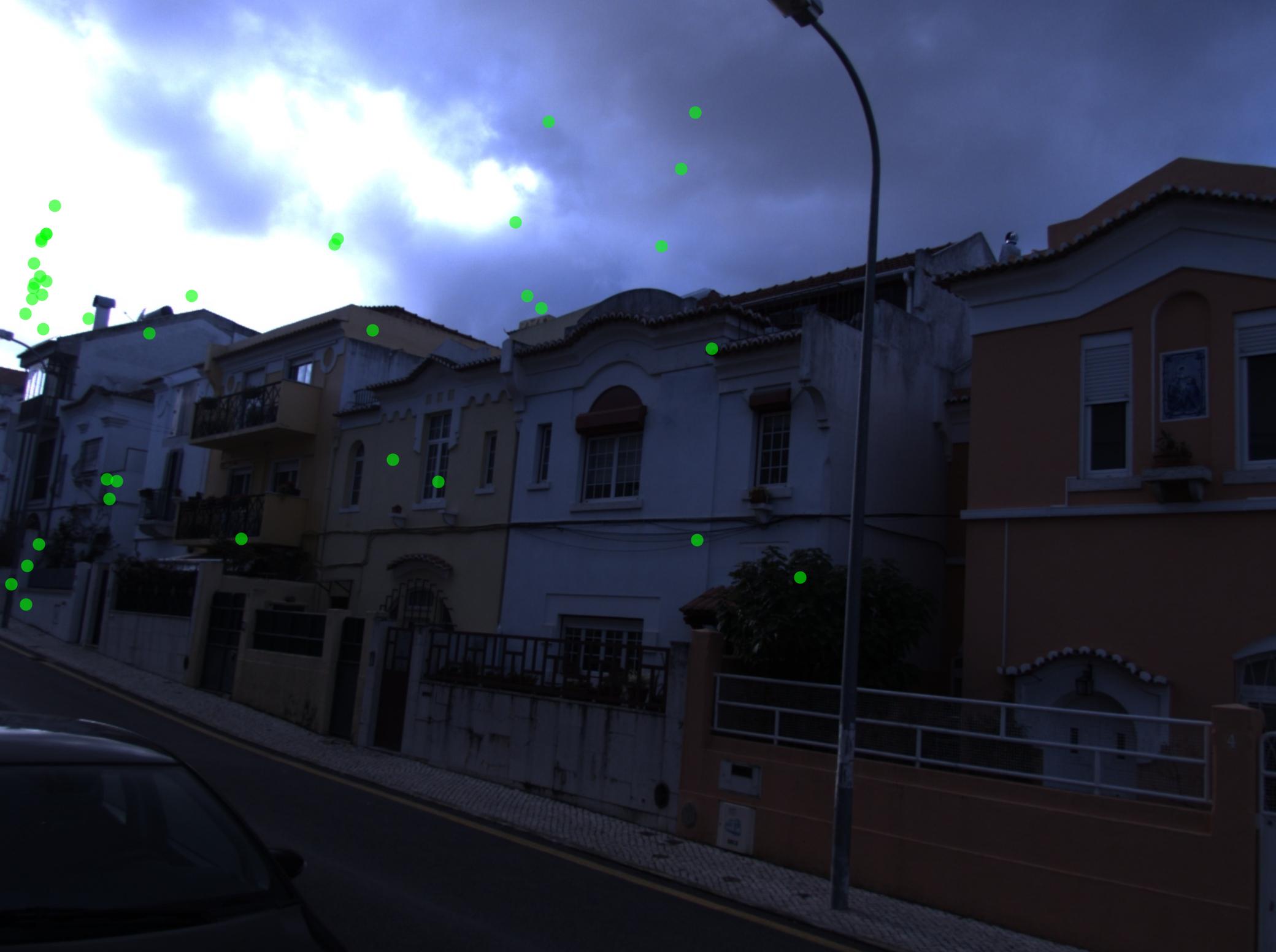}
    \includegraphics[height=0.053\textheight]{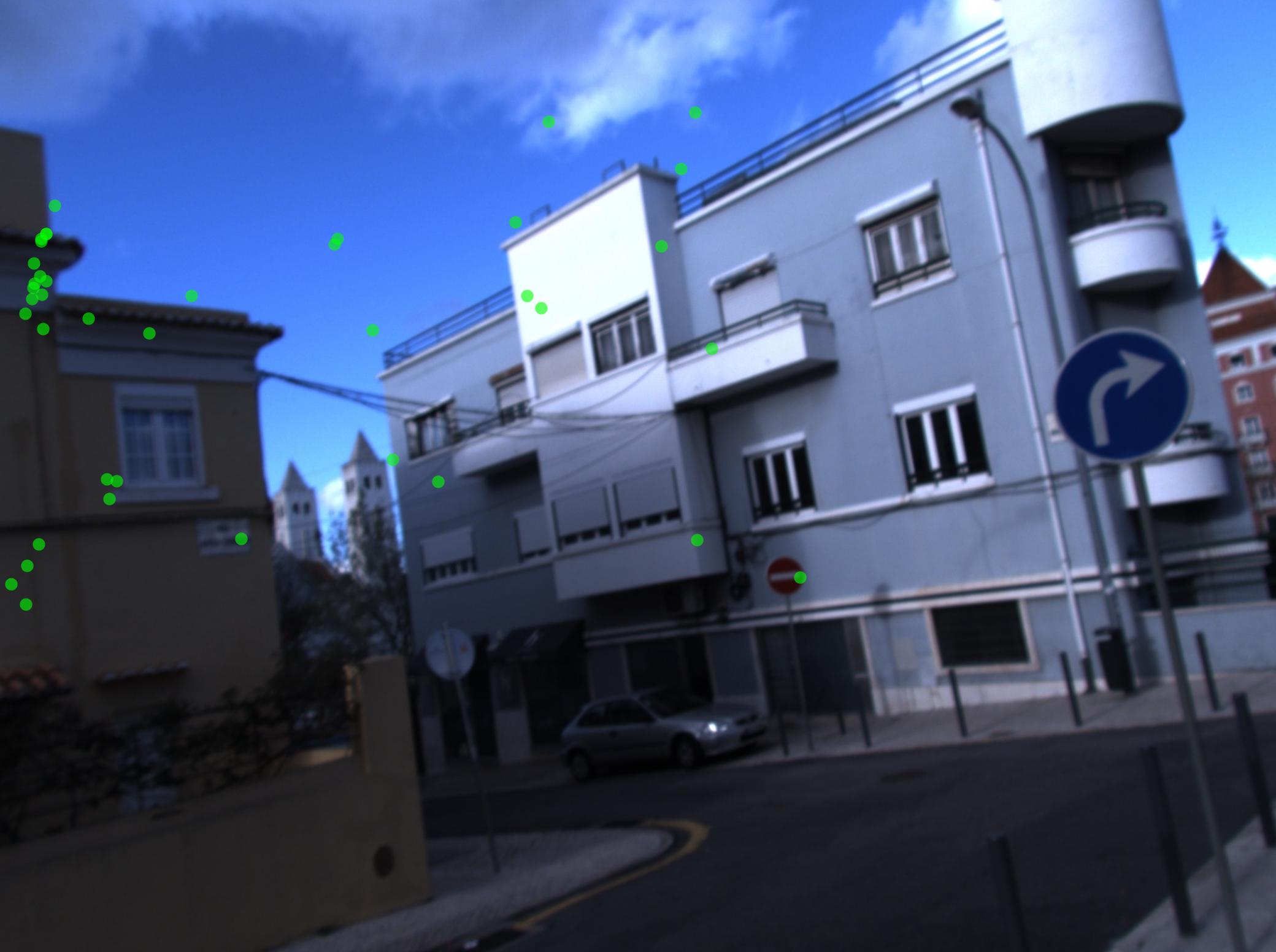}
    \label{fig:real2:images}}
    \end{tabular}
  \captionsetup[subfloat]{farskip=0pt,captionskip=0pt}
    \begin{tabular}[b]{c}%
    \subfloat[Recovered path using only Camera 1.]{\includegraphics[height=0.175\textheight]{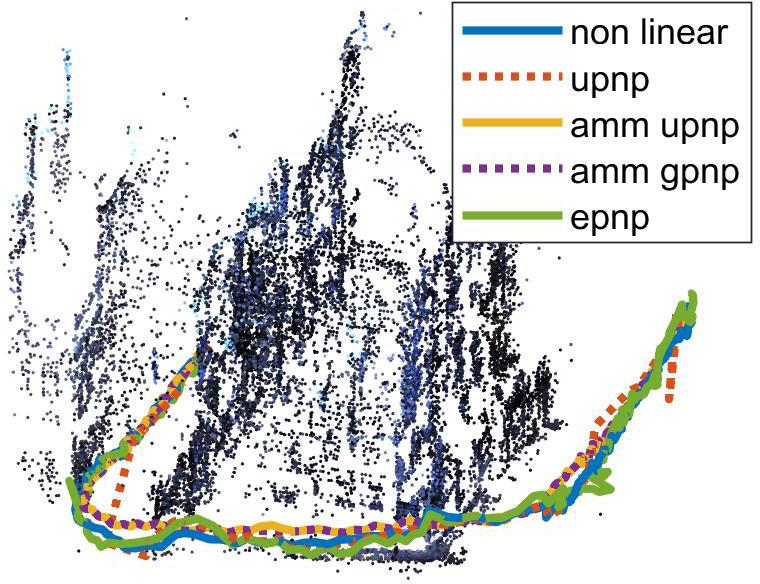}
    \label{fig:real2:results_b}
    }
    \subfloat[Recovered path using the three cameras]{\includegraphics[height=0.175\textheight]{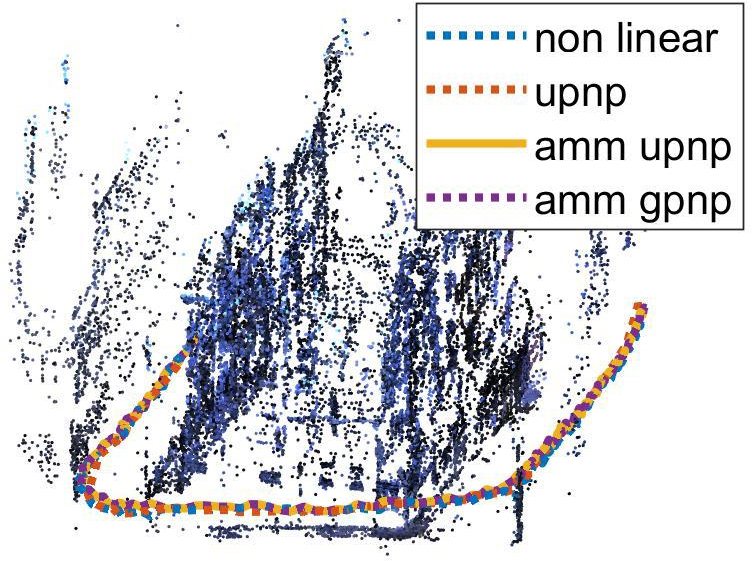}
    \label{fig:real2:results_c}
    }
    \end{tabular}
    }
    \caption{These figures show the recovered paths from the multi-perspective camera system shown in \protect\subref{fig:real2:setup}, whose set of images in a specific instance of time is given are shown in \protect\subref{fig:real2:images}. \protect\subref{fig:real2:results_b} shows the results obtained for the path using the absolute central case (only camera 1 was considered), and  \protect\subref{fig:real2:results_c} presents the results for the absolute noncentral case, where all the three cameras of the multi-perspective system were considered.}
    \label{fig:real2}
\end{figure*}

For the relative case (Fig.~\ref{fig:relative_pose_results_synthetic}), the rotation found is close to the {\tt ge} for each noise level and it takes significantly less time. The {\tt non linear} algorithm has the best accuracy, but its computation time is one order of magnitude higher than all other algorithms considered.

For the central absolute pose (Fig.~\ref{exp:abs_central_info}), the {\tt upnp} and {\tt non linear} present the same or higher accuracy than our method, but their computation time is one order of magnitude higher (10 times slower). The {\tt epnp} algorithm's computation time is similar to ours but significantly less accurate, while the minimal case with RANSAC ({\tt p3p (ransac)}) is slower and less accurate than ours.
For the general non-central absolute pose case (Fig.~\ref{exp:abs_noncentral_info}), concerning the {\tt upnp} and {\tt non linear}, the conclusion is the same as before, likewise the minimal case within the RANSAC framework {\tt gp3p (ransac)}.

From these results, one can conclude that the AMM framework proposed in this paper performs better than other methods, despite the fact that it involves an iterative set of simple optimization steps.

\subsection{Results with Real Data}
\label{exp:real_data}
For the experiments results with real data, we have considered a non-central multi-perspective imaging device, which is given by three perspective cameras with non-overlapped field of view (see the setup in Fig.~\ref{fig:real2}\subref{fig:real2:setup}). Datasets such as KITTI \cite{Geiger2012CVPR} usually consider stereo systems in which the cameras are aligned with the moving direction of the vehicle. In such cases, when we find the correspondences between two images, the projection lines associated with pixels corresponding to the projection of the same world point become nearly the same, making it difficult to recover the pose using the epipolar constraint (degenerate configuration). This new dataset was acquired to avoid degenerate configurations.

Images were acquired synchronously\footnote{We use the {\tt ROS} toolbox ({\tt http://www.ros.org/}) for that purpose.} from a walking path of around 200 meters (see examples of these images in Fig.~\ref{fig:real2}\subref{fig:real2:images}).
To get the 2D to 3D correspondences, we use the {\tt VisualSFM} framework \cite{Wu11,Wu13}.

Cameras' intrinsic parameters were computed offline.
The correspondences between image pixels that are the images of 3D points are converted into 3D projection lines by using the correspondent camera parameters and their transformation w.r.t. each other.
The bearing vectors (direction corresponding to the projection rays) and camera centers w.r.t the imaging coordinate system are given as input ($\mathcal{D}_i$) to the framework (as well as the 3D points), which were used to compute the absolute pose.

In this experiments, it was considered the following state-of-the-art methods: {\tt gpnp} presented in \cite{gerald08}; the {\tt upnp} \etal ~\cite{kneip14_1}; the miminimal solution {\tt gp3p} \cite{nister04_2}; and the {\tt non linear} method \cite{kneip14_1}. 
Looking at \ref{fig:real2}\subref{fig:real2:results_b} and \ref{fig:real2}\subref{fig:real2:results_c}, it is possible to conclude that all methods retrieve the path.

In terms of results, for the central case the following times were obtained: {\tt non linear} 18.00s; {\tt upnp} 0.75s; {\tt epnp} 0.07s; {\tt amm (gpnp)} 0.11s; and {\tt amm (upnp)} 0.11s (the values of time are the sum of the time computed along the path).
For the general mon-central case the following times were obtained: {\tt non linear} 37.59s; {\tt upnp} 1.14s; {\tt amm (gpnp)} 0.17s; and {\tt amm (upnp)} 0.27s.

These results are in accordance with the conclusions of the precedent subsection. Because of its simplicity, the proposed framework solves these problems faster than current state-of-the-art approaches designed to solve specific pose problems.

\section{Discussion}
\label{conclusions}
In this paper, we have proposed a general framework for solving pose problems. Instead of considering each one individually, we start from a general formulation of these kind of problems, and aimed at a framework for solving any pose problem. We state the problem as an optimization one, in which we use an alternating minimization strategy to relax the constraints associated with the non-linearities of the optimization function.

Theoretically, our framework comes with three different algorithms that were optimized for pose estimation purposes. As for inputs, in addition to the data, the proposed framework requires an objective function (which depends on the considered residuals and data) and their respective gradients w.r.t. the rotation and translation parameters, being therefore very easy to use because: 1) there is no need to eliminate unknown variables to relax the optimization process; and 2) no specific solvers are needed. The framework was included in the {\tt OpenGV} library, and will be made available for the community\footnote{Check the author's webpage.}.

In terms of experimental results, we run several tests using both synthetic and real data. The main conclusion is that, although the framework is general (in the sense that their solvers aim to solving any pose problem) and very easy to solve (requires few information on the used metric), the sensitivity to noise is not affected (note that this depends on the chosen residual formulation), while being considerably faster.

\bibliographystyle{IEEEtran}
\bibliography{./egbib.bib}

}

{

\title{\LARGE \bf
POSEAMM: A Unified Framework for Solving \\ Pose Problems using an Alternating Minimization Method \\
{\tt (Supplementary Material)}
}

\author{Jo\~{a}o Campos$^{1}$, Jo\~{a}o R. Cardoso$^{2}$, and Pedro Miraldo$^3$
\thanks{$^{1}$Jo\~{a}o Campos is with the ISR, Instituto Superior T\'{e}cnico, Univ. Lisboa, Portugal.
E-Mail:~{\tt\small jcampos@isr.tecnico.ulisboa.pt}.}%
\thanks{$^{2}$Jo\~{a}o Cardoso is with the ISEC, Instituto Politecnico de Coimbra, Portugal, and the Institute of Systems and Robotics, University of Coimbra, Portugal. E-Mail:~{\tt\small jocar@isec.pt}.}%
\thanks{$^{3}$P. Miraldo is with the KTH Royal Institute of Technology, Stockholm, Sweden.
E-Mail:~{\tt\small miraldo@kth.se}.}
}

\maketitle
\thispagestyle{empty}
\pagestyle{empty}

\begin{abstract}
This document contains the supplementary material of the paper entitled POSEAMM: A Unified Framework for Solving Pose Problems using an Alternating Minimization Method, that was accepted in the 2019 IEEE Int'l Conf. Robotics and Automation (ICRA). Here, we provide  the auxiliary calculations that led to the objective functions and their respective gradients (in terms of the rotation and translation parameters). In addition, we include the prototypes of the framework implemented in {\tt C++} in the {\tt OpenGV} toolbox, and show an example of its application as well. As an overview, in Sec.~\ref{app:general_relative_pose_sup_mat}, we will explain the calculations in Sec. III-A of the main paper. In Sec.~\ref{app:general_absolute_pose} and \ref{app:general_absolute_pose_upnp} we present the calculations for Secs. III-B and III-C in the main paper. In Sec.~\ref{app:example_code} we present the implementation of the objective functions.
\end{abstract}


\section{General Relative Pose Problem}
\label{app:general_relative_pose_sup_mat}
In this section we explain with detail how the objective function (Sec.~\ref{app:rgpnp_fo}) and their respective gradients (Sec.~\ref{app:rgpnp_grad}) presented in Eq. 9 and 10 of the main paper have been obtained.

\subsection{Objective function $\mathcal{F}(\mathbf{R},\mathbf{t})$}
\label{app:rgpnp_fo}

For the objective function, we considered the {\it Generalized Epipolar} constraint \cite{pless03}:
\begin{equation}
\label{sm:epipolar_constraint}
\mathbf{l}_{1}^{T}\mathbf{F}\,
\mathbf{l}_{2} = \mathbf{0}, \ \text{with} \ \mathbf{F} = \begin{bmatrix}
\mathbf{E} & \mathbf{R} \\
\mathbf{R} & \mathbf{0} \\
\end{bmatrix},
\end{equation}
where $\mathbf{l}_{1}$ and $\mathbf{l}_{2}$ are the \textit{Pl\"ucker} coordinates 
of two distinct 3D projection rays that intersect, in two distinct frame coordinates (let us say 1 and 2). $\mathbf{E}$ and $\mathbf{R}$ are the essential \cite{hartley01} and the rotation matrices that represent the transformation between the frame coordinates considered. Applying the Kronecker product (here denoted as $\otimes$) to
\eqref{sm:epipolar_constraint} yields
\begin{equation}
\label{sm:epipolar_constraint_linearized}
\left(\mathbf{l}_{2}^{T} \otimes \mathbf{l}_{1}^{T} \right)
\mathbf{f} = \mathbf{0},
\end{equation}
where $\mathbf{f}$ corresponds to $\mathbf{F}$ stacked column by column. Considering that some elements of the matrix $\mathbf{F}$ are zero, it is possible to verify that the entries in positions 22-24, 28-30, and 34-36 of the vector $\mathbf{f}$ are zero. Thus, we can rewrite \eqref{sm:epipolar_constraint_linearized} as:
\begin{equation}
    \mathbf{a}^{T} \mathbf{v} = 0 \ \ \text{ with }\ \ \mathbf{v} = \begin{bmatrix}\mathbf{e} \\ \mathbf{r}\end{bmatrix},
\end{equation}
in which $\mathbf{e}$ and $\mathbf{r}$ represent the essential and rotation matrices stacked. The vector $\mathbf{v}$ will have the same elements as $\mathbf{f}$ except for the ones that are null. The vector $\mathbf{a}$ can be obtained by eliminating the elements in positions 22-24, 28-30, and 34-36 of the vector $\left(\mathbf{l}_{2}^{T} \otimes \mathbf{l}_{1}^{T} \right) \in \mathbb{R}^{1\times 36}$, since they will multiply by $\mathbf{f}$'s null elements, and by taking the elements 4-6, 10-12 and 16-18 in $\left(\mathbf{l}_{2}^{T} \otimes \mathbf{l}_{1}^{T} \right)$ and summing them to the elements 19-21, 25-27 and 31-33, since they will be multiplying by the first, second and third column of $\mathbf{R}$, respectively. Now, each correspondence between $\mathbf{l}_1$ and $\mathbf{l}_2$ has a vector $\mathbf{a}_i$ associated. Thus, the objective function can be written as
\begin{multline}
\label{sm:of_rel_case}
\mathcal{F}(\mathbf{R}, \mathbf{t}) = \sum_{i=1}^{N}
\mathbf{v}^{T}\left(\mathbf{a}_{i} \mathbf{a}_{i}^{T} \right)\mathbf{v}\ \Rightarrow \\ \mathcal{F}(\mathbf{R}, \mathbf{t}) = \mathbf{v}^{T}\mathbf{M}\mathbf{v}, \text{where} \ \mathbf{M} = \sum_{i=1}^{N}\mathbf{a}_{i} \mathbf{a}_{i}^{T},
\end{multline}
yielding Eq. 9 of the paper.

\subsection{Gradients for $\nabla g\left(\mathbf{R}\right)$ and $\nabla h\left(\mathbf{t}\right)$}
\label{app:rgpnp_grad}
Here, we will give the full form of the rotation (here denoted as $\nabla g\left(\mathbf{R}\right)$) and the translation ($\nabla h\left(\mathbf{t}\right)$) gradients.

The essential matrix is given by $\mathbf{E} = \hat{\mathbf{t}}\mathbf{R}$, where $\hat{\mathbf{t}}$ is the skew-symmetric matrix associated with the translation vector. 
The explicit expression of $\mathbf{v}$ results from stacking the elements of $\mathbf{E}$:
\begin{equation}
 \mathbf{E} = \begin{bmatrix}
        t_2 r_3 - t_3 r_2 & t_2 r_6 - t_3 r_5 & t_2 r_9 - t_3 r_8 \\ 
     t_3 r_1 - t_1 r_3 & t_3 r_4 - t_1 r_6 & t_3 r_7 - t_1 r_9 \\ 
         t_1 r_2 - t_2 r_1 & t_1 r_5 - t_2 r_4 & t_1 r_8 - t_2 r_7 \\ 
         \end{bmatrix},
\end{equation}
followed by the stacking of the elements of $\mathbf{R}$. The vector $\mathbf{v}$ will depend on 12 distinct variables. Considering the objective function as given in \eqref{sm:of_rel_case} {\color{red}}, we have
\begin{equation}
    \mathcal{F}\left(\mathbf{R},\mathbf{t}\right) = \sum_{i=1}^{18} \sum_{j=1}^{18}\mathbf{v}_i \mathbf{M}_{ij} \mathbf{v}_j.
\end{equation}

Now, computing the derivative of $\mathcal{F}\left(\mathbf{R},\mathbf{t}\right)$ with respect to a variable $\lambda$ representing any element of the rotation matrix or the translation vector, and taking into account that $\mathbf{M}^{T} = \mathbf{M}$ yields
\begin{eqnarray}
    \frac{d\mathcal{F}\left(\mathbf{R},\mathbf{t}\right)}{d\lambda} & = & \sum_{i=1}^{18}\sum_{j=1}^{18} \left(\frac{d\mathbf{v}_i}{d\lambda} \mathbf{M}_{ij} \mathbf{v}_j +  \mathbf{v}_i \mathbf{M}_{ij} \frac{d\mathbf{v}_j}{d\lambda}\right)\nonumber  \\
    & = &
    \frac{d\mathbf{v}^{T}}{d\lambda} \mathbf{M} \mathbf{v} +  \mathbf{v}^{T} \mathbf{M} \frac{d\mathbf{v}}{d\lambda}  \label{sm:derivation_derivative}\\
   & = &
    \frac{d\mathbf{v}^{T}}{d\lambda} \mathbf{M} \mathbf{v} +  \frac{d\mathbf{v}^{T}}{d\lambda} \mathbf{M}^{T} \mathbf{v}  \nonumber\\
    & = &
    2\frac{d\mathbf{v}^{T}}{d\lambda} \mathbf{M} \mathbf{v}.\nonumber
\end{eqnarray}
Consider the derivatives of $\mathbf{v}$ in order to the three elements of the translation $\mathbf{t}$:
{ \setlength\arraycolsep{3pt}
\begin{align}
\frac{d\mathbf{v}^{T}}{d t_{1}} = &
    \begin{bmatrix}
    0 & -r_3 & r_2 & 0 & -r_6 & r_5 & 0 & -r_9 & r_8 & \mathbf{0}_{1 \times 9}
    \end{bmatrix}
    \\
    \frac{d\mathbf{v}^{T}}{d t_{2}} = &
    \begin{bmatrix}
    r_3 & 0 & -r_1 & r_6 & 0 & -r_4 & r_9 & 0 & -r_7 & \mathbf{0}_{1 \times 9}
    \end{bmatrix}
    \\
    \frac{d\mathbf{v}^{T}}{d t_{3}} = &
    \begin{bmatrix}
    -r_2 & r_1 & 0 & -r_5 & r_4 & 0 & -r_8 & r_7 & 0 & \mathbf{0}_{1 \times 9}
    \end{bmatrix}.
\end{align}
}Denoting the first, second and third columns of the rotation matrix by $\mathbf{r}_1$, $\mathbf{r}_2$ and $\mathbf{r}_3$, respectively, we can assemble the three previous equations as:
\begin{equation}
    \frac{d\mathbf{v}^{T}}{d\mathbf{t}} = 
    \begin{bmatrix}
    \hat{\mathbf{r}}_1 & \hat{\mathbf{r}}_2 & \hat{\mathbf{r}}_3 & \mathbf{0_{3 \times 9}}
    \end{bmatrix},
\end{equation}
where $\hat{\mathbf{r}}_i$ ($i=1,2,3$) stands to the skew-symmetric matrix associated to the vector ${\mathbf{r}}_i$.
Inserting this result in \eqref{sm:derivation_derivative} leads to the gradient of the translation given in the paper:
\begin{equation}
    \nabla h(\mathbf{t}) = 2 
    \frac{d\mathbf{v}^{T}}{d\mathbf{t}}\mathbf{M} \mathbf{v}.
\end{equation}

We proceed similarly to obtain the gradient of the rotation:
{ \setlength\arraycolsep{2.5pt}
\begin{align}
    \frac{d\mathbf{v}^{T}}{d r_{1}} = &
    \begin{bmatrix}
    0 & t_3 & -t_2 & \mathbf{0}_{1 \times 3} & \mathbf{0}_{1 \times 3} & 1 & 0 & 0 & \mathbf{0}_{1 \times 3} & \mathbf{0}_{1 \times 3}
    \end{bmatrix}
    \\
    \frac{d\mathbf{v}^{T}}{d r_{2}} = &
    \begin{bmatrix}
    -t_3 & 0 & t_1 & \mathbf{0}_{1 \times 3} & \mathbf{0}_{1 \times 3} & 0 & 1 & 0 & \mathbf{0}_{1 \times 3} & \mathbf{0}_{1 \times 3}
    \end{bmatrix}
    \\
    \frac{d\mathbf{v}^{T}}{d r_{3}} = &
    \begin{bmatrix}
    t_2 & -t_1 & 0 & \mathbf{0}_{1 \times 3} & \mathbf{0}_{1 \times 3} & 0 & 0 & 1 & \mathbf{0}_{1 \times 3} & \mathbf{0}_{1 \times 3}
    \end{bmatrix}
    \\
    \frac{d\mathbf{v}^{T}}{d r_{4}} = &
    \begin{bmatrix}
    \mathbf{0}_{1 \times 3} & 0 & t_3 & -t_2 & \mathbf{0}_{1 \times 3} & \mathbf{0}_{1 \times 3} & 1 & 0 & 0 & \mathbf{0}_{1 \times 3}
    \end{bmatrix}
    \\
    \frac{d\mathbf{v}^{T}}{d r_{5}} = &
    \begin{bmatrix}
    \mathbf{0}_{1 \times 3} & -t_3 & 0  & t_1 & \mathbf{0}_{1 \times 3} & \mathbf{0}_{1 \times 3} & 0 & 1 & 0 & \mathbf{0}_{1 \times 3}
    \end{bmatrix}
    \\
    \frac{d\mathbf{v}^{T}}{d r_{6}} = &
    \begin{bmatrix}
    \mathbf{0}_{1 \times 3} & t_2 & -t_1  & 0 & \mathbf{0}_{1 \times 3} & \mathbf{0}_{1 \times 3} & 0 & 0 & 1 & \mathbf{0}_{1 \times 3}
    \end{bmatrix}
    \\
    \frac{d\mathbf{v}^{T}}{d r_{7}} = &
    \begin{bmatrix}
    \mathbf{0}_{1 \times 3} & \mathbf{0}_{1 \times 3} & 0 & t_3 & -t_2 & \mathbf{0}_{1 \times 3} & \mathbf{0}_{1 \times 3} & 1 & 0 & 0
    \end{bmatrix}
    \\
    \frac{d\mathbf{v}^{T}}{d r_{8}} = &
    \begin{bmatrix}
    \mathbf{0}_{1 \times 3} & \mathbf{0}_{1 \times 3} & 0 & -t_3 & t_1 & \mathbf{0}_{1 \times 3} & \mathbf{0}_{1 \times 3} & 0 & 1 & 0
    \end{bmatrix}
    \\
    \frac{d\mathbf{v}^{T}}{d r_{9}} = &
    \begin{bmatrix}
    \mathbf{0}_{1 \times 3} & \mathbf{0}_{1 \times 3} & 0 & t_2 & -t_1 & \mathbf{0}_{1 \times 3} & \mathbf{0}_{1 \times 3} & 0 & 0 & 1
    \end{bmatrix}.
\end{align}}These nine equations contain the derivative of the objective function for each element of the rotation matrix. Writing them in a compact form will lead to:
{ \setlength\arraycolsep{4pt}
\begin{equation}
\frac{d\mathbf{v}^{T}}{d\mathbf{r}} =
    \begin{bmatrix}
    -\hat{\mathbf{t}} & \mathbf{0}_{3 \times 3} & \mathbf{0}_{3 \times 3} & \mathbf{I}_{3 \times 3} & \mathbf{0}_{3 \times 3} & \mathbf{0}_{3 \times 3} \\
    \mathbf{0}_{3 \times 3} & -\hat{\mathbf{t}} & \mathbf{0}_{3 \times 3} & \mathbf{0}_{3 \times 3} & \mathbf{I}_{3 \times 3} & \mathbf{0}_{3 \times 3} \\
    \mathbf{0}_{3 \times 3} & \mathbf{0}_{3 \times 3} & -\hat{\mathbf{t}} & \mathbf{0}_{3 \times 3} & \mathbf{0}_{3 \times 3} & \mathbf{I}_{3 \times 3} \\
    \end{bmatrix}.
\end{equation}}

Now, combining this result with \eqref{sm:derivation_derivative} gives the gradient of the translation:
\begin{equation}
    \nabla g(\mathbf{R}) = 2 
    \frac{d\mathbf{v}^{T}}{d\mathbf{r}}\mathbf{M} \mathbf{v}.
\end{equation}

\section{Objective function for the General Absolute Pose}
\label{app:general_absolute_pose}
Here we address the general absolute pose (Sec.~III-B of the main paper), by considering the geometric distance between  3D points and an inverse 3D projection ray as a metric for the objective function.

\subsection{Objective function $\mathcal{F}(\mathbf{R},\mathbf{t})$}
\label{app:gpnp_fo}
The objective function is based on the geometric distance derived in \cite{gerald08}:
{\begin{multline}
    \label{sm:absolute_parcel}
    \mathcal{F}(\mathbf{R}, \mathbf{t}) = 
    \sum_{i=1}^{N} \mathbf{e}^T \mathbf{e}, \ \text{where} \ 
    \mathbf{e} =  \left(\mathbf{I} - \mathbf{V}_i\right)
    \left(\mathbf{R} \mathbf{x}_{i} + \mathbf{t} - \mathbf{c}_{i}\right), \\ \text{and}\
\mathbf{V}_i = \frac{\mathbf{v}_{i}\mathbf{v}_{i}^{T}}{\mathbf{v}_{i}^{T} \mathbf{v}_{i}}.
\end{multline}}The vector $\mathbf{x}_i$ corresponds to the $i^{th}$ 3D point in the world frame, $\mathbf{c}_{i}$ corresponds to the $i^{th}$ camera's position and $\mathbf{v}_i$ corresponds to the projection onto ray direction.

Despite not being necessary in our framework, it would be advantageous, for efficiency purposes, to rewrite the above expression in matricial form. 
For convenience, we define
\begin{equation}
\label{sm:Q_value}
\mathbf{w}_i = \mathbf{R} \mathbf{x}_{i}  - \mathbf{c}_{i} \ \ \text{and} \ \
\mathbf{Q}_i = \left(\mathbf{I} - \mathbf{V}_i\right)^{T}\left(\mathbf{I} - \mathbf{V}_i\right),
\end{equation}
where  $\mathbf{Q}_i$ is a symmetric matrix. Replacing these expressions in \eqref{sm:absolute_parcel} we obtain
\begin{equation}
\label{obj_func_first}
    \mathcal{F}(\mathbf{R}, \mathbf{t}) = 
    \sum_{i=1}^{N}\mathbf{t}^{T} \mathbf{Q}_i \mathbf{t} +
    \mathbf{t}^{T}\sum_{i=1}^{N}\left(-2 \mathbf{Q}_i \mathbf{w}_i \right)  + 
    \sum_{i=1}^{N}\mathbf{w}_i^{T}\mathbf{Q}_i\mathbf{w}_i.
\end{equation}

The dependence of the objective function on the translation is exhibited in the first term.
A linear term in the translation $\mathbf{t}$ and a crossed term in $\mathbf{t}$ and $\mathbf{R}$ is displayed in the second term.
The third term gives rise to a quadratic term in $\mathbf{R}$, a linear term in $\mathbf{R}$ and a constant term in the final expression.
Applying the Kronecker product to the second term leads to the following:

\begin{eqnarray}
\mathbf{t}^{T}\sum_{i=1}^{N}\left(-2 \mathbf{Q}_i \mathbf{w}_i \right) = \mathbf{t}^{T}\sum_{i=1}^{N}-2 \mathbf{Q}_i \left(\mathbf{R} \mathbf{x}_{i}  - \mathbf{c}_{i}\right) \\
    = \mathbf{t}^{T}\sum_{i=1}^{N}-2\mathbf{Q}_i\mathbf{R}\mathbf{x}_i + 
    \mathbf{t}^{T}\sum_{i=1}^{N}2\mathbf{Q}_i\mathbf{c}_{i}\\
    = \mathbf{t}^{T}\left[-2\sum_{i=1}^{N}\left(\mathbf{x}_i^{T} \otimes \mathbf{Q}_i\right)\right]\mathbf{r} + \mathbf{t}^{T}\sum_{i=1}^{N}2\mathbf{Q}_i\mathbf{c}_{i}\nonumber \\
     = \mathbf{t}^{T}\left[-2\sum_{i=1}^{N}\left(\mathbf{x}_i^{T} \otimes \mathbf{Q}_i\right)\right]\mathbf{r} + \left[\sum_{i=1}^{N}2\mathbf{c}_{i}^{T}\mathbf{Q}_i\right]\mathbf{t}.
     \label{sm:second_term_final}
\end{eqnarray}
Now, we obtain two more terms that will appear in the final expression of the objective function. The final terms will be obtained by expanding the last term in \eqref{obj_func_first}. Replacing $\mathbf{Q}_i$ by its value \eqref{sm:Q_value}, we obtain:

\begin{equation}
\begin{split}
\sum_{i=1}^{N}\mathbf{w}_i^{T}\mathbf{Q}_i\mathbf{w}_i = \sum_{i=1}^{N} \mathbf{x}_i^{T} \mathbf{R}^{T} \mathbf{Q}_i \mathbf{R} \mathbf{x}_i +  \sum_{i=1}^{N} -2 \mathbf{c}_i^{T} \mathbf{Q}_i \mathbf{R} \mathbf{x}_i + \\ \sum_{i=1}^{N} \mathbf{c}_i \mathbf{Q}_i \mathbf{c}_i  =  \\ \sum_{i=1}^{N} \mathbf{x}_i^{T} \mathbf{R}^{T} \left(\mathbf{I} - \mathbf{V}_i\right)^{T} \left(\mathbf{I} - \mathbf{V}_i\right)\mathbf{R} \mathbf{x}_i + \sum_{i=1}^{N} -2 \mathbf{c}_i^{T} \mathbf{Q}_i \mathbf{R} \mathbf{x}_i +\\ \sum_{i=1}^{N} \mathbf{c}_i \mathbf{Q}_i \mathbf{c}_i.
\end{split}
\label{sm:gpnp_third_label}
\end{equation}

We note that, for a specific correspondence, the first element of the sum in \eqref{sm:gpnp_third_label} can be seen as the scalar product of a vector $\mathbf{b}_i = \left(\mathbf{I} - \mathbf{V}_i\right)\mathbf{R} \mathbf{x}_i$ by itself. Applying the Kronecker product to the vector $\mathbf{b}$ and to the last equation's second term in \eqref{sm:gpnp_third_label} leads to
\begin{equation}
\label{sm:b_and_second}
\begin{split}
    \mathbf{b}_i = \left[\mathbf{x}_i^{T} \otimes \left(\mathbf{I} - \mathbf{V}_i\right)\right]\mathbf{r}\ \text{ and }\\
    \sum_{i=1}^{N}-2\mathbf{c}_i^{T}\mathbf{Q}_i\mathbf{R} \mathbf{x}_i = \left[\sum_{i=1}^{N}-2\mathbf{x}_i^{T} \otimes \left(\mathbf{c}^{T}\mathbf{Q}_i\right)\right]\mathbf{r}.
\end{split}
\end{equation}
Now, replacing \eqref{sm:b_and_second} in \eqref{sm:gpnp_third_label}, we obtain an expression for the third term in \eqref{obj_func_first}:
\begin{equation}
\label{sm:final_third_term}
\begin{split}
    \sum_{i=1}^{N}\mathbf{w}_i^{T}\mathbf{Q}_i\mathbf{w}_i = \\ 
    \mathbf{r}^{T} \left[ \sum_{i=1}^{N}\left[\mathbf{x}_i^{T} \otimes \left(\mathbf{I} - \mathbf{V}_i\right)\right]^{T}\left[\mathbf{x}_i^{T} \otimes \left(\mathbf{I} - \mathbf{V}_i\right)\right] \right]\mathbf{r} + \\
    \left[\sum_{i=1}^{N}-2\mathbf{x}_i^{T} \otimes \left(\mathbf{c}^{T}\mathbf{Q}_i\right)\right]\mathbf{r} +
    \sum_{i=1}^{N} \mathbf{c}_i \mathbf{Q}_i \mathbf{c}_i.
\end{split}
\end{equation}

Inserting \eqref{sm:second_term_final} and \eqref{sm:final_third_term} in \eqref{obj_func_first} gives rise to the following expression for the objective function in a matrix form, that can be easily used to calculate the gradients: 

\begin{equation}
\label{sm:final_objective_fun}
\begin{split}
    \mathcal{F}(\mathbf{R}, \mathbf{t}) = 
    \mathbf{r}^{T} \underbrace{\left[ \sum_{i=1}^{N}\left[\mathbf{x}_i^{T} \otimes \mathbf{Q}_i\right]^{T}\left[\mathbf{x}_i^{T} \otimes \mathbf{Q}_i\right] \right]}_{\mathbf{M}_{rr}}\mathbf{r} + \\
    \underbrace{\left[\sum_{i=1}^{N}-2\mathbf{x}_i^{T} \otimes \left(\mathbf{c}^{T}\mathbf{Q}_i\right)\right]}_{\mathbf{v}_{r}^T}\mathbf{r} + \\
    \mathbf{t}^{T}\underbrace{\left[-2\sum_{i=1}^{N}\left(\mathbf{x}_i^{T} \otimes \mathbf{Q}_i\right)\right]}_{\mathbf{M}_{tr}}\mathbf{r} + \\
    \mathbf{t}^{T} \underbrace{\left[\sum_{i=1}^{N}\mathbf{Q}_i\right]}_{\mathbf{M}_{tt}}\mathbf{t} + 
    \underbrace{\left[\sum_{i=1}^{N}2\mathbf{c}_{i}^{T}\mathbf{Q}_i\right]}_{\mathbf{v}_{t}^T}\mathbf{t} +
    \underbrace{\sum_{i=1}^{N} \mathbf{c}_i \mathbf{Q}_i \mathbf{c}_i}_{c},
\end{split}
\end{equation}
which corresponds to Eq. 11 of the main paper. Given that $\left|\mathbf{v}_i\right| = 1$, $\mathbf{Q}_i = \mathbf{I} - \mathbf{V}_i$. 

\subsection{Gradients for $\nabla g\left(\mathbf{R}\right)$ and $\nabla h\left(\mathbf{t}\right)$}
The Euclidean gradients of $\nabla g\left(\mathbf{R}\right)$ and $\nabla h\left(\mathbf{t}\right)$ (Eqs. 12 and 13 of the main paper) can be computed directly from \eqref{sm:final_objective_fun}, by applying well known results of matrix computations \cite{lutkepohl97}:
\begin{equation}
\begin{split}
    \nabla g(\mathbf{R}) = 2  \mathbf{M}_{rr}\mathbf{r} + \mathbf{v}_r + \mathbf{M}_{tr}^{T} \mathbf{t} \ \ \text{ and } \ \ 
    \nabla h(\mathbf{t}) = 2 \mathbf{M}_{tt} \mathbf{t} + \\ \mathbf{M}_{tr} \mathbf{r} + \mathbf{v}_t.
\end{split}
\end{equation}

\section{Objective Function with the UPnP Residual}
\label{app:general_absolute_pose_upnp}
In this section, we explain how the objective function in the last application example  (Sec. III-C of the main paper) and its gradients have been obtained.

\subsection{Objective function $\mathcal{F}(\mathbf{R},\mathbf{t})$}
\label{app:upnp_fo}
To get an expression to the objective function, we proceed similarly as in \cite{kneip14_1}. The starting point are the equation already presented in (Sec.~III-C):
\begin{equation}
\label{app:upnp_1st_system_2}
    \alpha_i \mathbf{v}_i + \mathbf{c}_i = \mathbf{R} \mathbf{p}_i + \mathbf{t}, \ \ i=1,\ldots,N.
\end{equation}
which can be written in the form
\begin{equation}
    \begin{split}
    \underbrace{\begin{bmatrix}
        \mathbf{v}_1 & \mathbf{0} & \cdots & \mathbf{0} & -\mathbf{I} \\
        \mathbf{0} & \mathbf{v}_2 & \cdots & \mathbf{0} & -\mathbf{I} \\
        \vdots & \vdots & \ddots & \vdots & \vdots \\
        \mathbf{0} & \mathbf{0} & \cdots & \mathbf{v}_n & - \mathbf{I}
    \end{bmatrix}}_{\mathbf{A}}
    \underbrace{\begin{bmatrix}
    \alpha_1 \\
    \alpha_2 \\
    \vdots      \\
    \alpha_n \\
    \mathbf{t}
    \end{bmatrix}}_{\mathbf{x}} = \\
    \underbrace{\begin{bmatrix} 
    \mathbf{R} & \mathbf{0} & \cdots & \mathbf{0} \\
    \mathbf{0} & \mathbf{R} & \cdots & \mathbf{0} \\
        \vdots & \vdots & \ddots & \vdots \\
    \mathbf{0} & \mathbf{0} & \cdots & \mathbf{R}  
    \end{bmatrix}}_{\mathbf{W}}
    \underbrace{\begin{bmatrix} 
    \mathbf{p}_1 \\
    \mathbf{p}_2 \\
    \vdots       \\
    \mathbf{p}_n
    \end{bmatrix}}_{\mathbf{b}} - 
    \underbrace{\begin{bmatrix} 
    \mathbf{c}_1 \\
    \mathbf{c}_2 \\
    \vdots       \\
    \mathbf{c}_n
    \end{bmatrix}}_{\mathbf{w}},
    \end{split}
\end{equation}
or, more compactly, as:
\begin{equation}
\label{sm:stacked_constraints}
\mathbf{A} \mathbf{x} = \mathbf{W} \mathbf{b} - \mathbf{w}\ \Leftrightarrow \
 \mathbf{x} = \left(\mathbf{A}^{T}\mathbf{A}\right)^{-1} \mathbf{A}^{T} \left(\mathbf{W} \mathbf{b} - \mathbf{w}\right) .
\end{equation}

As in \cite{kneip14_1}, we write \eqref{sm:stacked_constraints} as
\begin{equation}
 \mathbf{x} = \begin{bmatrix} \mathbf{U} \\ \mathbf{V} \end{bmatrix} \left(\mathbf{W} \mathbf{b} - \mathbf{w}\right).
\end{equation}
Note that the dimensions of the matrix $\left(\mathbf{A}\mathbf{A}^{T}\right)^{-1}\mathbf{A}^{T}$ are $\left(N+3\right) \times 3N$, meaning that there are matrices $\mathbf{U}_{N\times 3N}$ and $\mathbf{V}_{3 \times 3N}$ such that
\begin{equation}
\left(\mathbf{A}\mathbf{A}^{T}\right)^{-1}\mathbf{A}^{T} = \begin{bmatrix}\mathbf{U} \\ \mathbf{V} \end{bmatrix}
\end{equation}
and, as a consequence,
\begin{equation}
    \begin{bmatrix}
    \alpha_1 \\
    \alpha_2 \\
    \vdots      \\
    \alpha_n \\
    \end{bmatrix} = \mathbf{U} \left(\mathbf{W}\mathbf{b} - \mathbf{w}\right)
    \ \ \text{ and } \ \
    \mathbf{t} = \mathbf{V} \left(\mathbf{W}\mathbf{b} - \mathbf{w}\right).
\end{equation}

Contrarily to \cite{kneip14_1}, we are only interested in eliminating the dependence on the depth's ($\alpha_i$). Therefore, from the previous expressions,  
\begin{equation}
\label{sm:first_alpha}
\alpha_i = \mathbf{u}_i^{T} \left(\mathbf{W} \mathbf{b} - \mathbf{w}\right), 
\end{equation}
where $\mathbf{u}_i^{T}$ is the $i^{th}$ row of $\mathbf{U}$. Since equation \eqref{sm:first_alpha} does not depend explicitly on the data nor on the parameter we are interested in ($\mathbf{R}$ and $\mathbf{t}$), we consider the $1\times 3$ vector $\mathbf{u}_{ij}$ that corresponds to the $j^{th}$ 3 element-vector of $\mathbf{u}_i$. By making use of these vectors, \eqref{sm:first_alpha} becomes:
\begin{equation}
\label{sm:second_alpha}
\alpha_i = \sum_{j=1}^{N} \mathbf{u}_{ij}^{T} \left(\mathbf{R}\mathbf{p}_j - \mathbf{c}_j\right)\ \Leftrightarrow \
\alpha_i = \sum_{j=1}^{N} \mathbf{u}_{ij}^{T} \mathbf{R}\mathbf{p}_j - \sum_{j=1}^{N} \mathbf{u}_{ij}^{T} \mathbf{c}_j .
\end{equation}

From \eqref{app:upnp_1st_system_2} and \eqref{sm:second_alpha}, and using Kronecker products, we have:
\begin{equation}
\label{sm:final_result}
\begin{split}
    \eta_i \left(\mathbf{R}, \mathbf{t}\right) = \alpha_i \mathbf{v}_i + \mathbf{c}_i - \mathbf{R}\mathbf{p}_i - \mathbf{t} \nonumber \\
    =  \mathbf{v}_i\left(\sum_{j=1}^{N} \mathbf{u}_{ij}^{T} \mathbf{R}\mathbf{p}_j \right) - \left(\sum_{j=1}^{N} \mathbf{u}_{ij}^{T} \mathbf{c}_j \right) \mathbf{v}_i +  \\ \mathbf{c}_i - \mathbf{R}\mathbf{p}_i - \mathbf{t} \nonumber \\
    = \mathbf{v}_i \left(\sum_{j=1}^{N} \mathbf{u}_{ij}^{T} \mathbf{R}\mathbf{p}_j \right) - \left(\sum_{j=1}^{N} \mathbf{u}_{ij}^{T} \mathbf{c}_j \right)\mathbf{v}_i + \\ \mathbf{c}_i - \left(\mathbf{p}_i^{T} \otimes \mathbf{I}\right)\mathbf{r} - \mathbf{t}      \\
    = \left[\sum_{j=1}^{N} \mathbf{v}_i\left(\mathbf{p}_j^{T} \otimes \mathbf{u}_{ij}^{T}\right) \right] \mathbf{r} - \left(\sum_{j=1}^{N} \mathbf{u}_{ij}^{T} \mathbf{c}_j \right)\mathbf{v}_i  + \\ \mathbf{c}_i - \left(\mathbf{p}_i^{T} \otimes \mathbf{I}\right)\mathbf{r} - \mathbf{t} \nonumber
    \\
    = \left(\left[\sum_{j=1}^{N} \mathbf{v}_i\left(\mathbf{p}_j^{T} \otimes \mathbf{u}_{ij}^{T}\right) \right] - \left(\mathbf{p}_i^{T} \otimes \mathbf{I}\right)\right)\mathbf{r} - \\ \left(\sum_{j=1}^{N} \mathbf{u}_{ij}^{T} \mathbf{c}_j \right)\mathbf{v}_i  + \mathbf{c}_i  - \mathbf{t}.\nonumber
\end{split}
\end{equation}

By considering the objective function as being
\begin{equation}
    \mathcal{F}\left(\mathbf{R}, \mathbf{t}\right) = \sum_i^{N}\eta_i^{T}\left(\mathbf{R},\mathbf{t}\right)\eta_i\left(\mathbf{R},\mathbf{t}\right),
\end{equation}
and using the residuals in \eqref{sm:final_result}, we get the objective function shown in Eq. 11 of the main paper, where:


\begin{align}
    \label{sm:final_matrices_gpnp}
    \begin{split}
    \mathbf{M}_{rr} = \sum_{i=1}^{N}\left(\left[\sum_{j=1}^{N} \mathbf{v}_i\left(\mathbf{p}_j^{T} \otimes \mathbf{u}_{ij}^{T}\right) \right] - \left(\mathbf{p}_i^{T} \otimes \mathbf{I}\right)\right)^{T}  \\ \left(\left[\sum_{j=1}^{N} \mathbf{v}_i\left(\mathbf{p}_j^{T} \otimes \mathbf{u}_{ij}^{T}\right) \right] - \left(\mathbf{p}_i^{T} \otimes \mathbf{I}\right)\right) \\
    \mathbf{v}_{r} = \sum_{i=1}^{N}2\left(\left[\sum_{j=1}^{N} \mathbf{v}_i\left(\mathbf{p}_j^{T} \otimes \mathbf{u}_{ij}^{T}\right) \right] - \left(\mathbf{p}_i^{T} \otimes \mathbf{I}\right)\right)^{T} \\
    \left[\mathbf{c}_i - \left(\sum_{j=1}^{N} \mathbf{u}_{ij}^{T} \mathbf{c}_j \right)\mathbf{v}_i \right] \\
    \mathbf{M}_{tr} = \sum_{i=1}^{N}-2\left(\left[\sum_{j=1}^{N} \mathbf{v}_i\left(\mathbf{p}_j^{T} \otimes \mathbf{u}_{ij}^{T}\right) \right] - \left(\mathbf{p}_i^{T} \otimes \mathbf{I}\right)\right)\\
    \mathbf{M}_{tt} = \sum_{i=1}^{N} \mathbf{I}\\
    \mathbf{v}_{t} = \sum_{i=1}^{N}-2\left[\mathbf{c}_i - \left(\sum_{j=1}^{N} \mathbf{u}_{ij}^{T} \mathbf{c}_j \right)\mathbf{v}_i \right] \\
    c = \sum_{i=1}^{N}\left[\mathbf{c}_i - \left(\sum_{j=1}^{N} \mathbf{u}_{ij}^{T} \mathbf{c}_j \right)\mathbf{v}_i \right]^{T} \\ \left[\mathbf{c}_i - \left(\sum_{j=1}^{N} \mathbf{u}_{ij}^{T} \mathbf{c}_j \right)\mathbf{v}_i\right],
    \end{split}
\end{align}

\subsection{Gradients for $\nabla g\left(\mathbf{R}\right)$ and $\nabla h\left(\mathbf{t}\right)$}
As before, the computation of these euclidean gradients $\nabla g\left(\mathbf{R}\right)$ and $\nabla h\left(\mathbf{t}\right)$ are computed directly from \eqref{sm:final_objective_fun}:
\begin{equation}
    \nabla g(\mathbf{R}) = 2  \mathbf{M}_{rr}\mathbf{r} + \mathbf{v}_r + \mathbf{M}_{tr}^{T} \mathbf{t} \ \ \text{ and } \ \ 
    \nabla h(\mathbf{t}) = 2 \mathbf{M}_{tt} \mathbf{t} + \mathbf{M}_{tr} \mathbf{r} + \mathbf{v}_t,
\end{equation}
where $\mathbf{M}_{rr}$, $\mathbf{M}_{tt}$, $\mathbf{M}_{tr}$, $\mathbf{v}_{r}$, $\mathbf{v}_{t}$, and $c$ are presented in \eqref{sm:final_matrices_gpnp}.

\section{Use our Framework: An example of its applications}
\label{app:example_code}
\subsection{Code Prototype}

In this section we present the Pure Abstract Class in {\tt C++}. Whatever class is used to build the objective function, it must provide an implementation for the functions below. These are passed on to the AMM solver that will handle the problem.
\begin{lstlisting}
#ifndef OBJECTIVEFUNCTIONINFO_H
#define OBJECTIVEFUNCTIONINFO_H

#include <Eigen/Dense>
#include <opengv/types.hpp>


class ObjectiveFunctionInfo{
public:

virtual double objective_function_value(
                            const opengv::rotation_t & rotation,
                            const opengv::translation_t & translation
                            ) = 0;
virtual opengv::rotation_t rotation_gradient(
                            const opengv::rotation_t & rotation,
                            const opengv::translation_t & translation
                            ) = 0;
virtual opengv::translation_t translation_gradient(
                            const opengv::rotation_t & rotation,
                            const opengv::translation_t & translation
                            ) = 0;
};

#endif

}
\end{lstlisting}

\subsection{Implementation of the objective and gradients functions}
In this section we show the code where the matrices $\mathbf{M}_{rr}$, $\mathbf{v}_{r}$, $\mathbf{M}_{tr}$, $\mathbf{M}_{tt}$, $\mathbf{v}_{t}$ and $c$ is created for our algorithm {\tt amm (gpnp)}. In the implementation the constant is not considered. This is an example of how easy it is to solve a problem using our framework.
\begin{lstlisting}
GlobalPnPFunctionInfo::GlobalPnPFunctionInfo(
    const opengv::absolute_pose::AbsoluteAdapterBase & adapter
    ){

  opengv::Indices indices(adapter.getNumberCorrespondences());
  int total_points = (int) indices.size();
 
  Mt =  Eigen::Matrix3d::Zero(3,3);
  Mrt = Eigen::MatrixXd::Zero(3,9);
  Mr =  Eigen::MatrixXd::Zero(9,9);
  vt =  Eigen::VectorXd::Zero(3,1);
  vr =  Eigen::VectorXd::Zero(9,1);
  
  Eigen::Matrix3d id = Eigen::Matrix3d::Identity(3,3);

  Eigen::Matrix<double,3,1> vi;
  Eigen::Matrix<double,3,1> xi;
  Eigen::Matrix<double,3,1> ci;
  Eigen::Matrix3d Qi;
  Eigen::Matrix3d Vi;
  Eigen::MatrixXd Mr_i  = Eigen::MatrixXd::Zero(3,9);
  Eigen::MatrixXd Mrt_i = Eigen::MatrixXd::Zero(3,9);
  Eigen::Matrix3d I_V   = Eigen::MatrixXd::Zero(3,3);
  Eigen::MatrixXd vr_1  = Eigen::MatrixXd::Zero(1,3);
  Eigen::MatrixXd vr_2  = Eigen::MatrixXd::Zero(1,9);
  for( int i = 0; i < total_points; i++ )
  {
    vi = adapter.getCamRotation(indices[i]) * adapter.getBearingVector(indices[i]);
    xi = adapter.getPoint(indices[i]);
    ci = adapter.getCamOffset(indices[i]);
    Vi = vi * vi.transpose() / (vi.transpose() * vi);
    Qi = (id - Vi).transpose() * (id - Vi);
    
    I_V = id - Vi;
    Mt = Mt + Qi;
    vt = vt - (2 * Qi * ci);
    //Constant = Constant + ci.transpose() * Qi * ci;
   
    //Calculate Mr  
    Mr_i.block<3,3>(0,0) = xi(0,0) * I_V;
    Mr_i.block<3,3>(0,3) = xi(1,0) * I_V;
    Mr_i.block<3,3>(0,6) = xi(2,0) * I_V;
   
    Mr = Mr + Mr_i.transpose() * Mr_i;
   
    //Calculate Mrt
    Mrt_i.block<3,3>(0,0) = xi(0,0) * Qi;
    Mrt_i.block<3,3>(0,3) = xi(1,0) * Qi;
    Mrt_i.block<3,3>(0,6) = xi(2,0) * Qi;
    
    Mrt = Mrt + 2*Mrt_i;
       
    //Calculate vr
    vr_1 = ci.transpose() * Qi;
    vr_2.block<1,3>(0,0) = -2 * xi(0,0) * vr_1;
    vr_2.block<1,3>(0,3) = -2 * xi(1,0) * vr_1;
    vr_2.block<1,3>(0,6) = -2 * xi(2,0) * vr_1;
    vr = vr + vr_2.transpose();
  }
}
\end{lstlisting}

The objective function value is implemented simply as:
\begin{lstlisting}
double GlobalPnPFunctionInfo::objective_function_value(
    const opengv::rotation_t & rotation, const opengv::translation_t & translation
    ){
    
  const double * p = &rotation(0);
  Map<const Matrix<double,1,9> > r(p, 1, 9);
  Eigen::MatrixXd e = (translation.transpose() * Mt * translation)
                    + (translation.transpose() * Mrt *  r.transpose())
                    + (vt.transpose() * translation)
                    + (r * Mr * r.transpose() )
                    + (vr.transpose() * r.transpose());
                    
  return ( e(0,0));
}

\end{lstlisting}
The rotation gradient $\nabla g\left(\mathbf{R}\right)$:
\begin{lstlisting}
opengv::rotation_t GlobalPnPFunctionInfo::rotation_gradient(
    const opengv::rotation_t & rotation, const opengv::translation_t & translation
    ){
    
    const double * p = &rotation(0);
    Map<const Matrix<double,1,9> > r(p, 1, 9);
    Eigen::MatrixXd result = (2 * Mr * r.transpose()) + 
                                ( Mrt.transpose() * translation ) + vr;
    double * ptr = &result(0);
    Map<Matrix<double, 3,3> > m(ptr, 3, 3);
  
  return m;
}
\end{lstlisting}
The translation gradient $\nabla h\left(\mathbf{R}\right)$:
\begin{lstlisting}
opengv::translation_t GlobalPnPFunctionInfo::translation_gradient(
    const opengv::rotation_t & rotation, const opengv::translation_t & translation
    ){
    
    const double * p = &rotation(0);
    Map<const Matrix<double,1,9> > r(p, 1, 9);
    
    return ( (2 * Mt * translation) + (  Mrt * r.transpose() ) + vt );
}

\end{lstlisting}

}

\end{document}